\documentclass{article}

\usepackage[table]{xcolor}
\usepackage[final]{corl_2026}
\usepackage{url}
\usepackage{booktabs}
\usepackage{tabularx}
\usepackage{amsfonts}
\usepackage{nicefrac}
\usepackage{microtype}
\usepackage{multirow,graphicx}

\usepackage{float}

\usepackage{amsmath}
\usepackage{pifont}
\usepackage{subcaption}
\usepackage{makecell}
\usepackage{marvosym}
\usepackage[ruled,vlined,linesnumbered]{algorithm2e}
\usepackage{amssymb}
\usepackage{bm}
\SetKwComment{Comment}{\textcolor{gray}{$\triangleright$}\ }{}
\newcommand{\mycomment}[1]{\Comment*[r]{\textcolor{gray}{#1}}}
\definecolor{checkgreen}{RGB}{34,139,34}
\definecolor{xmarkred}{RGB}{200,40,40}

\newcommand{\cmarkc}{\textcolor{checkgreen}{\ding{51}}}
\newcommand{\xmarkc}{\textcolor{xmarkred}{\ding{55}}}

\title{SafeDojo: Safe Reinforcement Learning for VLA via Interactive World Model}

\hypersetup{
  pdftitle={SafeDojo: Safe Reinforcement Learning for VLA via Interactive World Model},
  pdfauthor={Kai Tang, Peidong Jia, Zhong Chu, Jiajun Cao, Jixian Wu, Bohua Zhang, Rui Ma, Fangyuan Zhao, Sixiang Chen, Yichen Guo, Xiaowei Chi, Chun-Kai Fan, Kevin Zhang, Jinchang Xu, Fubing Yang, Weishi Mi, Xiaozhu Ju, Jian Tang, Shanghang Zhang},
  pdfsubject={Safe Reinforcement Learning for Vision-Language-Action Models},
  pdfkeywords={Safe Reinforcement Learning, Vision-Language-Action, World Model}
}

\author{
  \normalfont
  \vspace{1.0em}
  \begin{tabular}{c}
  \normalsize Kai Tang\textsuperscript{1,2,*}\quad
  Peidong Jia\textsuperscript{1,2,*,\textdagger}\quad
  Zhong Chu\textsuperscript{1,3,*}\quad
  Jiajun Cao\textsuperscript{1,2,*}\\[-0.05em]
  \normalsize Jixian Wu\textsuperscript{1}\quad
  Bohua Zhang\textsuperscript{1,5}\quad
  Rui Ma\textsuperscript{1}\quad
  Fangyuan Zhao\textsuperscript{1}\quad
  Sixiang Chen\textsuperscript{1}\\[-0.05em]
  \normalsize Yichen Guo\textsuperscript{1,3}\quad
  Xiaowei Chi\textsuperscript{1,4}\quad
  Chun-Kai Fan\textsuperscript{1,2}\quad
  Kevin Zhang\textsuperscript{1}\\[-0.05em]
  \normalsize Jinchang Xu\textsuperscript{1,2}\quad
  Fubing Yang\textsuperscript{5}\quad
  Weishi Mi\textsuperscript{2}\\[-0.12em]
  \normalsize Xiaozhu Ju\textsuperscript{2}\quad
  Jian Tang\textsuperscript{2,\Letter}\quad
  Shanghang Zhang\textsuperscript{1,\Letter}\\[0.55em]
  \small \textsuperscript{1}State Key Laboratory of Multimedia Information Processing,\\[-0.08em]
  \small School of Computer Science, Peking University\\[-0.08em]
  \small \textsuperscript{2}Beijing Innovation Center of Humanoid Robotics\\[-0.08em]
  \small \textsuperscript{3}Nanyang Technological University\\[-0.08em]
  \small \textsuperscript{4}Hong Kong University of Science and Technology\\[-0.08em]
  \small \textsuperscript{5}University of Electronic Science and Technology of China\\[0.35em]
  \small \textsuperscript{*}Equal contribution. \textsuperscript{\textdagger}Project Leader. \textsuperscript{\Letter}Corresponding authors.\\
  \small \Letter\enspace\href{mailto:shanghang@pku.edu.cn}{\texttt{shanghang@pku.edu.cn}}
  \end{tabular}
}

\begin{document}
\maketitle
\vspace{-0.38in}

\noindent\makebox[\textwidth][c]{%
  \small\textbf{Project page: }
  \href{https://kai030113.github.io/safedojo/}
  {\texttt{kai030113.github.io/safedojo}}%
}
\par\vspace{-0.05in}

\begin{abstract}
Safe control is a prerequisite for real-world embodied intelligence, for which safe reinforcement learning has emerged as a promising paradigm. However, existing safe reinforcement learning methods either require costly real-world exploration or depend on hand-crafted safety functions. Neither scales to vision-language-action models deployed in open-world physical environments. We propose \textbf{SafeDojo}, a model-based safe reinforcement learning framework that improves vision-language-action policies through world-model imagination. SafeDojo performs online reinforcement learning on top of an interactive video world model. A ResNet success classifier scores predicted future frames, while a lightweight safety head predicts per-step contact probabilities from the current latent context and proposed action chunk, jointly assessing task execution and trajectory safety. These decoupled signals are balanced through Lagrangian-constrained GRPO. On SafeLIBERO Level I, SafeDojo improves average safe-success rate by 8.25 percentage points over the strongest baseline; on the zero-shot Level II setting, it is competitive with the strongest model-free baseline. Real-world Franka deployment further shows the best average task and safe-success rates across five tasks. These results support world-model-based safe reinforcement learning as a promising path toward safer embodied intelligence.

\end{abstract}
\vspace{-0.4cm}
\begin{figure}[H]
    \centering    \includegraphics[width=0.99\linewidth]{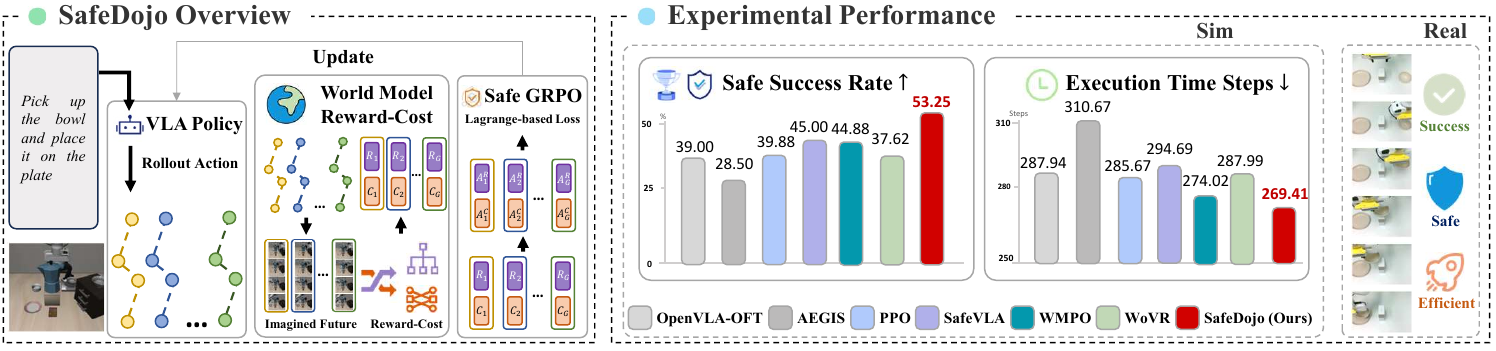}
    \caption{\textbf{Overview of SafeDojo.} SafeDojo enhances VLA policies with world-model-based reward--cost evaluation and safe GRPO, boosting safe success and efficiency in simulation and real-world scenarios. The performance panels compare against all baselines; complete comparisons are reported in Tables~\ref{tab:main_level1} and~\ref{tab:main_level2}.}
    \label{fig:teaser}
    \vspace{-0.4cm}
\end{figure}

\section{Introduction}
Safe control is an urgent challenge for deploying 
autonomous robots in real-world environments~\cite{brunke2022safe, gu2024review}. Among 
various safety concerns, collision avoidance during contact-rich robotic manipulation stands out as one of the most 
critical, since even a single violation can cause  task failure, irreversible 
hardware damage or environmental disruption~\cite{khatib1986real, haddadin2008collision, haddadin2009requirements, haddadin2017robot, zanchettin2016safety, lasota2017survey, ding2022safety}. As 
vision-language-action (VLA) models become the dominant 
paradigm for general-purpose robotic manipulation~\cite{brohan2023rt2,openx2023rtx,kim2024openvla,octo2024,black2024pi0}, we 
focus on a key open problem, \textbf{\textit{learning 
collision-free safe VLA policies in obstacle-rich environments}}. 
Unlike low-dimensional control with explicit or compactly learned safety constraints~\cite{ames2019control,huang2024safedreamer}, VLA policies must infer safety boundaries from high-dimensional visual observations across diverse and previously unseen obstacle configurations~\cite{zhang2025safevla}.
This challenge is compounded by the irreversibility of 
physical violations, demanding that unsafe actions be 
anticipated and prevented before execution.
Existing approaches to safe robotic manipulation span 
several paradigms, including inference-time control barrier 
functions~\cite{hu2025vlsa}, model-free constrained 
reinforcement learning (RL)~\cite{zhang2025safevla}, and 
model-based safe RL~\cite{huang2024safedreamer, cao2025fosp}. 
Control barrier functions provide theoretical safety 
guarantees but require precise geometric modeling and 
camera calibration that are impractical for open-world 
VLA deployment. Model-free methods enforce safety through 
penalty-based optimization but cannot anticipate violations 
before they occur. Model-based safe RL addresses this limitation by performing anticipatory rollouts in imagination. However, existing methods largely operate in low-dimensional state spaces with simulator-provided cost functions, neither of which directly transfers to VLA models. Extending model-based safe 
reinforcement learning to VLA is particularly challenging 
because task progress and collision avoidance are 
inherently conflicting objectives, and safety signals 
must be learned from high-dimensional visual observations 
rather than analytically defined. To our knowledge, model-based safe reinforcement learning 
for vision-language-action models remains unexplored. 
We propose SafeDojo as the first framework to address it.

We propose SafeDojo, to our knowledge the first world model-based safe RL framework for VLA models. SafeDojo is motivated by \textbf{a central challenge in safe VLA learning, where policy improvement requires exploration, yet a single unsafe contact in the real world can cause task failure, hardware damage, or environmental disruption.} SafeDojo performs online policy optimization entirely within an interactive video world model, avoiding risky real-world exploration. The world model rolls out candidate trajectories into action-conditioned future video frames and latent representations, from which a dual-branch evaluator estimates dense task reward and safety cost signals. Specifically, a finetuned ResNet success classifier predicts per-step task progress from imagined frames, while a lightweight but effective safety head estimates per-step safety-cost signals from the current latent context and the proposed action chunk. These dense signals are aggregated into trajectory-level reward-cost pairs and optimized with a Lagrangian-based safety-constrained GRPO objective. This design enables SafeDojo to improve task success and collision avoidance under explicit safety constraints, without hand-crafted safety functions or risky real-world trials.

Our contributions are summarized as follows. 
\textbf{(1) World Model based Reward-Cost.}
We develop an interactive video world model for safe VLA rollouts, together with a dual-branch evaluator that estimates decoupled task reward and safety cost from imagined dynamics.
\textbf{(2) Safety-Constrained RL.}
We introduce Lagrangian-based safety-constrained GRPO to optimize policies under explicit safety constraints, avoiding fixed penalty while coordinating task success and collision avoidance. \textbf{(3) Superior Empirical Performance.} SafeDojo achieves strong gains on SafeLIBERO and real-world manipulation, improving the Level I average safe-success rate by $8.25$ percentage points over the strongest baseline while achieving the best average task success and execution efficiency on SafeLIBERO.
\vspace{-0.3cm}
\section{Related Work}
\vspace{-0.3cm}
Existing VLA safety mechanisms rely on perception augmentation~\cite{yu2026safenight,son2026thermoact}, inference-time intervention~\cite{hu2025vlsa,zhai2026cofreevla,gu2025safe}, or policy-level alignment~\cite{zhang2025safevla,zhang2024grape}, but do not optimize VLA policies with model-based safe RL. 
Safe RL under Constrained Markov Decision Processes (CMDPs)~\cite{altman1999constrained} offers a principled formulation, yet model-free methods~\cite{achiam2017constrained,ray2019benchmarking,tessler2019rcpo,zhang2020focops,stooke2020responsive} require costly and unsafe exploration, while model-based methods~\cite{thomas2021safe,huang2024safedreamer,hogewind2023safeslac,cao2025fosp,nakamura2025latent,seo2025unisafe} mostly target generic control with predefined scalar costs. 
World-model-based VLA post-training~\cite{zhu2026wmpo,jiang2026wovr,xiao2025worldenv,li2025vlarft,liu2026worldvlaloop} optimizes task return in imagination but lacks safety constraints. 
Detailed related work is provided in Appendix~\ref{app:related_work}.
\vspace{-0.4cm}
\section{Problem Definition and Preliminaries}
\vspace{-0.2cm}
\subsection{Problem Formulation}
\label{sec:problem}

\paragraph{Extending VLA MDPs to CMDPs.}
Standard VLA can be modeled as a Markov Decision Process (MDP) 
$\mathcal{M} = (\mathcal{S}, \mathcal{A}, P, r, \rho_0, \gamma)$, 
where $s_t\in\mathcal{S}$ denotes vision-language states, 
$a_t\in\mathcal{A}$ is the action space, and 
$r:\mathcal{S}\times\mathcal{A}\rightarrow\{0,1\}$ is the task reward indicator. $P$, $\rho_0$, and $\gamma$ follow standard MDP notation. 
We extend this formulation to a Constrained MDP (CMDP)~\cite{altman1999constrained} 
$\mathcal{M}^c = (\mathcal{S}, \mathcal{A}, P, r, c, \rho_0, \gamma)$ 
by introducing a safety cost 
$c:\mathcal{S}\times\mathcal{A}\rightarrow[0,1]$, requiring the VLA policy $\pi_\theta$ to maximize 
$J^r(\pi_\theta)=\mathbb{E}_{\pi_\theta}[\sum_t\gamma^t r_t]$ 
subject to 
$J^c(\pi_\theta)=\mathbb{E}_{\pi_\theta}[\sum_t\gamma^t c_t]\le d$, where $d$ denotes the allowable safety budget.

\paragraph{Model-based Safe RL for VLA.}
Under this CMDP formulation, model-based safe reinforcement learning for VLA improves the policy using imagined trajectory feedback, rather than relying on static safety rules, post-hoc action filtering, or risky real-world exploration.
Each policy rollout induces a trajectory $\tau$ with task rewards and safety costs accumulated over time. 
These accumulated signals are converted into advantages that drive policy-gradient updates of the form 
$\nabla_\theta \log \pi_\theta(a_t|s_t)A_t$, where task rewards encourage goal-directed actions and safety costs suppress collision-prone ones.
$\tau$ is considered safe-successful only if it completes the instructed task without incurring collision cost, making safe-success rate a stricter criterion than task success alone. 

\subsection{Preliminaries}
\vspace{-0.2cm}

Our method builds on Group Relative Policy Optimization (GRPO), which optimizes a policy using relative feedback among multiple sampled trajectories without learning a separate value critic. 
For a given state, the old policy $\pi_\theta^{\mathrm{old}}$ samples a group of $G$ trajectories $\{\tau_i\}_{i=1}^{G}$ and obtains scalar rewards $\{R_i\}_{i=1}^{G}$. 
GRPO normalizes rewards within the group to obtain $\hat{A}_i=(R_i-\mu_G(R))/(\sigma_G(R)+\epsilon)$ and updates the policy with the clipped objective:
\begin{equation}
\label{eq:grpo}
\mathcal{L}_{\mathrm{GRPO}}
=
-
\mathbb{E}_{i,t}
\left[
\min
\left(
\rho_{i,t}(\theta)\hat{A}_i,
\mathrm{clip}
\left(
\rho_{i,t}(\theta),1-\epsilon,1+\epsilon
\right)
\hat{A}_i
\right)
\right],
\end{equation}
where $\rho_{i,t}(\theta)=\pi_\theta(a_{i,t}|s_{i,t})/\pi_\theta^{\mathrm{old}}(a_{i,t}|s_{i,t})$ is the probability ratio. 
This formulation reinforces above-average samples and suppresses below-average ones, making it suitable for VLA optimization.
\section{Method}
\vspace{-0.2cm}

\begin{figure}[t]
    \centering
    \includegraphics[width=0.98\linewidth]{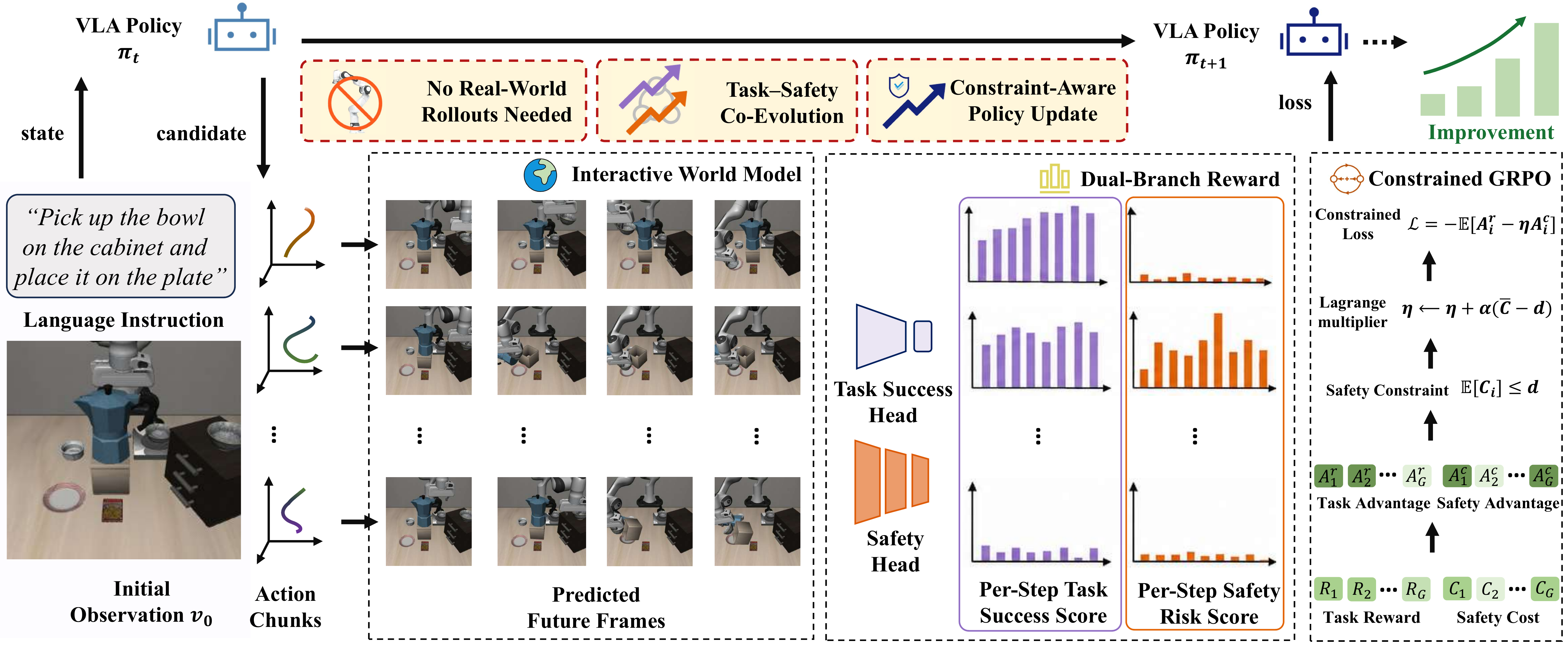}
    \caption{\textbf{Detailed SafeDojo Pipeline.} SafeDojo optimizes VLA policies entirely inside an interactive video world model by rolling out candidate action trajectories into imagined future dynamics. Task reward and safety cost are decoupled and optimized via Lagrangian-based constrained GRPO, improving task success while reducing safety risks without potentially damaging real-world rollouts.}
    \label{fig:method}
    \vspace{-0.5cm}
\end{figure}

Building on the safe VLA policy optimization problem defined in Sec.~\ref{sec:problem}, we propose \textbf{SafeDojo}, to our knowledge the first model-based safe reinforcement learning framework for VLA policy improvement. SafeDojo is motivated by the core tension between policy exploration for task improvement and unsafe contacts that may cause task failure, hardware damage, or environmental disruption. As illustrated in Fig.~\ref{fig:method}, SafeDojo addresses this challenge through three key innovations, including world model based virtual safety exploration in Sec.~\ref{sec:world_model}, decoupled reward-cost estimation in Sec.~\ref{sec:reward_cost}, and Lagrangian-based safety-constrained GRPO in Sec.~\ref{sec:cgrpo}. Together, these designs enable VLA policies to improve task success while reducing collision risks, without damage-prone real-world trials.

\subsection{Chunk-wise Interactive World Model Rollouts}
\label{sec:world_model}

Instead of collecting policy rollouts in the physical environment, SafeDojo performs model-in-the-loop rollouts inside an interactive world model $p_\phi$. 
We instantiate the interactive world model as an action-conditioned video predictor that generates future visual observations and latent dynamics. 
Given an initial visual observation $o_0$ and a language instruction $l$, the current VLA policy $\pi_\theta$ samples a group of candidate trajectories $\{\tau_i\}_{i=0}^{G-1}$, where each trajectory consists of $K$ action chunks. For an $H$-step action chunk $a_{i,k,1:H}$, the world model predicts $H$ temporally aligned future frames and latent states:
\begin{equation}
\{(\hat{o}_{i,k,h}, \hat{z}_{i,k,h})\}_{h=1}^{H}
=p_\phi(\hat{o}_{i,k,0}, z^{\mathrm{ctx}}_{i,k}, a_{i,k,1:H}, l),
\quad k=0,\ldots,K-1.
\label{eq:world_model_rollout}
\end{equation}
Here, $z^{\mathrm{ctx}}_{i,k}$ is the latent context available before rolling out chunk $k$. We build $p_\phi$ on Wan 2.2~\cite{wan2025} with cross-attention action conditioning.
Each action chunk consists of $H=8$ control steps, where each step is a $7$-DoF command (a $6$-DoF delta end-effector pose plus a gripper action) following the OpenVLA-OFT action representation and normalized with the suite-specific statistics.

However, chunk-wise prediction introduces a train-test mismatch between clean demonstration observations used during training and self-generated observations used during inference.
This mismatch becomes more severe in long-horizon rollouts due to exposure error, where the world model repeatedly conditions on its own imperfect predictions and accumulates errors across chunks.
To improve rollout robustness, we adopt a static-video augmentation during world-model training that occasionally freezes the conditioning frames, exposing $p_\phi$ to degenerate temporal contexts.
This augmentation makes the model more tolerant to imperfect intermediate predictions during autoregressive chunk-wise rollout.

\begin{algorithm}[t]
\caption{SafeDojo: Safe RL for VLA via Interactive World Model}
\label{alg:safedojo_short}
\KwIn{
VLA policy $\pi_{\theta}$, reference policy $\pi_{\theta}^{\mathrm{ref}}$, 
world model $p_{\phi}$, task reward head $f_{\mathrm{task}}$, safety cost head $f_{\mathrm{safe}}$, 
allowable budget $d$, group size $G$, multiplier $\eta$, multiplier update rate $\alpha_{\eta}$.
}
\KwOut{Optimized $\pi_{\theta}^{*}$ for task success under safety constraints.}

Initialize $\pi_{\theta}^{\mathrm{old}}\leftarrow \pi_{\theta}$, $\eta\geq0$, $d=0.2$, $\alpha_\eta=0.05$, $G=16$\;

\While{not converged}{
    Sample $(l,o_0)$\;

    $\{\tau_i\}_{i=1}^{G}\sim\pi_{\theta}^{\mathrm{old}}(\cdot\mid l,o_0)$
    \mycomment{candidate trajectories}

    \For{$i=1,\ldots,G$}{
        $\hat{\tau}_i \leftarrow p_{\phi}(o_0,l,\tau_i)$
         \mycomment{rollout, Eq.\,\ref{eq:world_model_rollout}}

        $\{r_{i,k,h},c_{i,k,h}\}_{k=0,h=1}^{K-1,H}\leftarrow 
        (f_{\mathrm{task}},f_{\mathrm{safe}})(\hat{\tau}_i)$
        \mycomment{dense scores, Eq.\,\ref{eq:per_step_reward_cost}}

        $(r_i,c_i)\leftarrow \mathrm{Agg}(\{r_{i,k,h},c_{i,k,h}\}_{k,h})$
        \mycomment{aggregation, Eq.\,\ref{eq:trajectory_reward_cost}}
    }

    $\hat{A}_i^r\leftarrow\mathrm{Norm}_{G}(r_i)$, 
    $\hat{A}_i^c\leftarrow\mathrm{Norm}_{G}(c_i)$
    \mycomment{advantages, Eq.\,\ref{eq:reward_cost_advantage}}
    $\eta \uparrow \text{ if } \frac{1}{G}\sum_{i=1}^{G} c_i>d,\ \text{else } \eta \downarrow$ \mycomment{multiplier, Eq.\,\ref{eq:lagrange_update}}
    $\hat{A}_i^{\mathrm{safe}}\leftarrow \hat{A}_i^r-\eta\hat{A}_i^c$
    \mycomment{safe advantage, Eq.\,\ref{eq:safe_advantage}}

    $\theta\leftarrow\arg\min_{\theta}\mathcal{L}_{\mathrm{Safe\text{-}GRPO}}(\theta)$
    \mycomment{policy update, Eq.\,\ref{eq:safe_grpo_loss}}

    $\pi_{\theta}^{\mathrm{old}}\leftarrow\pi_{\theta}$\;
}

\Return $\pi_{\theta}^{*}$\;
\end{algorithm}
\vspace{-0.3cm}

\subsection{Decoupled Task Reward and Safety Cost Estimation}
\label{sec:reward_cost}

Given an imagined trajectory $\hat{\tau}_i=\{(\hat{o}_{i,k,h},\hat{z}_{i,k,h},a_{i,k,h})\}_{k=0,h=1}^{K-1,H}$, SafeDojo evaluates it with decoupled task-reward and safety-cost signals. 
This decoupling is essential for safe VLA optimization, since task progress and safety risk are not always aligned. A trajectory may make progress toward the instruction while causing unsafe contact, or remain collision-free but fail to complete the task. 
Therefore, instead of collapsing task execution and safety into a hand-crafted scalar reward, SafeDojo estimates task reward and safety cost separately, providing structured feedback for subsequent constrained policy optimization.

The success head scores each predicted frame; the safety head uses the current latent context and proposed action chunk to predict $H$ contact probabilities:
\begin{equation}
r_{i,k,h}=f_{\mathrm{task}}(\hat{o}_{i,k,h},l),
\qquad
c_{i,k,h}=\left[f_{\mathrm{safe}}(z^{\mathrm{ctx}}_{i,k},a_{i,k,1:H})\right]_h,
\label{eq:per_step_reward_cost}
\end{equation}
Both outputs are aligned at the control-step level, although their inputs differ. We aggregate them into trajectory-level task reward and safety cost:
\begin{equation}
r_i
=
\frac{1}{KH}
\sum_{k=0}^{K-1}\sum_{h=1}^{H}r_{i,k,h},
\qquad
c_i
=
\frac{1}{M}
\sum_{c\in\mathrm{TopM}(\{c_{i,k,h}\}_{k,h})}
c,
\label{eq:trajectory_reward_cost}
\end{equation}
where $M$ denotes the number of high-risk steps used for safety aggregation. We use this top-$M$ mean predicted cost as a surrogate for the discounted cumulative CMDP cost, optimizing $\mathbb{E}[c_i]\le d$ in imagination.
In implementation, we set $M=16$. The task reward branch $f_{\mathrm{task}}$ is instantiated with a ResNet success classifier finetuned on rollout frames with per-frame success labels, while the safety cost branch $f_{\mathrm{safe}}$ is a compact convolutional latent encoder followed by an action-conditioned MLP head that outputs a per-step obstacle-contact probability over the proposed action chunk.

The resulting pair $(r_i, c_i)$ provides a reward-cost characterization of each candidate trajectory. A trajectory with high task reward and low safety cost corresponds to successful and safe execution, while a high-cost trajectory is discouraged even if it appears task-progressive. Conversely, a low-cost but low-reward trajectory is safe but ineffective. This decoupled evaluation allows SafeDojo to distinguish these cases and optimize task success and safety through an explicit constrained objective.

\subsection{Lagrangian-based Safety Constrained GRPO}
\label{sec:cgrpo}

A straightforward way to apply GRPO for safe VLA learning is reward shaping, which forms a scalar reward $\tilde{r}_i=r_i-\lambda c_i$ and then applies the vanilla GRPO objective in Eq.~\ref{eq:grpo}. 
This strategy treats safety as a fixed penalty rather than an explicit constraint, making the reward-cost trade-off sensitive to the manually chosen weight $\lambda$. More importantly, it does not explicitly enforce a safety boundary, allowing high-reward trajectories to remain favored even when they violate the desired cost budget.

Inspired by Constrained GRPO~\cite{girgis2026constrainedgrpo}, SafeDojo instead decouples task reward and safety cost, and optimizes them with a Lagrangian-based safety-constrained GRPO objective. Given the reward-cost pair $(r_i,c_i)$ estimated in Sec.~\ref{sec:reward_cost}, we compute group-relative reward and cost advantages separately:
\begin{equation}
\hat{A}_i^r
=
\frac{
r_i-\mu_G(r)
}{
\sigma_G(r)+\epsilon
},
\qquad
\hat{A}_i^c
=
\frac{
c_i-\mu_G(c)
}{
\sigma_G(c)+\epsilon
},
\label{eq:reward_cost_advantage}
\end{equation}
where $\mu_G(\cdot)$ and $\sigma_G(\cdot)$ denote the mean and standard deviation over the $G$ candidate trajectories sampled for the same instruction-observation context. 
Unlike reward shaping, which normalizes a pre-combined scalar reward, this separate normalization preserves the relative structure of task reward and safety cost before Lagrangian combination.

To optimize this surrogate-constrained objective, SafeDojo combines the separately normalized reward and cost advantages using a non-negative Lagrange multiplier $\eta$:
\begin{equation}
\hat{A}_i^{\mathrm{safe}}
=
\hat{A}_i^r
-
\eta \hat{A}_i^c .
\label{eq:safe_advantage}
\end{equation}
$\eta$ is then updated according to the batch-level constraint violation:
\begin{equation}
\eta
\leftarrow
\left[
\eta
+
\alpha_\eta
\left(
\frac{1}{G}\sum_{i=1}^{G} c_i - d
\right)
\right]_+,
\label{eq:lagrange_update}
\end{equation}
where $d$ is the budget on the top-$M$ surrogate cost in Eq.~\ref{eq:trajectory_reward_cost}, $\alpha_\eta$ is the multiplier learning rate, and $[\cdot]_+$ projects $\eta$ to be non-negative.
When the average safety cost exceeds $d$, $\eta$ increases to penalize unsafe trajectories more strongly. Otherwise, it decreases to allow more task-oriented improvement. We set $d=0.2$ and $\alpha_\eta=0.05$.

We obtain the final safety-constrained GRPO objective by replacing the vanilla advantage $\hat{A}_i$ in Eq.~\ref{eq:grpo} with $\hat{A}_i^{\mathrm{safe}}$ and adding KL regularization to the reference policy:
\begin{equation}
\mathcal{L}_{\mathrm{Safe\text{-}GRPO}}
=
-
\mathbb{E}_{i,t}
\left[
\min
\left(
\rho_{i,t}(\theta)\hat{A}_i^{\mathrm{safe}},
\mathrm{clip}
\left(
\rho_{i,t}(\theta),1-\epsilon,1+\epsilon
\right)
\hat{A}_i^{\mathrm{safe}}
\right)
\right]
+
\beta D_{\mathrm{KL}}
\left(
\pi_\theta \Vert \pi_\theta^{\mathrm{ref}}
\right),
\label{eq:safe_grpo_loss}
\end{equation}
where $\pi_\theta^{\mathrm{ref}}$ is the reference policy and $\beta$ controls the regularization strength. In experiments, we set $\beta=0$ and rely on the GRPO clipping ratio for the trust-region constraint. Algorithm~\ref{alg:safedojo_short} summarizes training, including world model rollouts, reward-cost estimation, and safety-constrained GRPO.

\section{Experiments}

\subsection{Experiments Setup}

\noindent\textbf{Experimental Details.} We evaluate SafeDojo on SafeLIBERO~\cite{hu2025vlsa}, a safety-critical benchmark built on LIBERO~\cite{liu2023libero} in Robosuite~\cite{zhu2020robosuite} with a single-arm Franka Emika Panda 7-DoF manipulator. 
SafeLIBERO includes four suites, Spatial, Goal, Object, and Long, each with four tasks under two obstacle-interference levels. 
Level I places obstacles near the target object, while Level II places them farther away but along the motion path. We train on Level I only and evaluate on both levels to test generalization, and further validate SafeDojo on 5 real-world Franka tasks.

\noindent\textbf{Baselines.} We compare SafeDojo with six baselines across four categories, all built on the same OpenVLA-OFT~\cite{kim2025fine} variant with parallel discrete action-token prediction, and initialized from a shared SFT checkpoint obtained by fine-tuning OpenVLA-OFT on collision-free SafeLIBERO demonstrations.
This variant enables token-level log-probability computation for GRPO.
OpenVLA-OFT serves as our Safe-SFT baseline: it is supervised-finetuned on the collision-free expert demonstrations, without RL or an additional safety mechanism.
AEGIS~\cite{hu2025vlsa} represents inference-time safety by applying a plug-and-play CBF-based action filter.
For model-free RL, we include reward-only PPO and SafeVLA~\cite{zhang2025safevla}, which formulates safety under the CMDP framework; both use the same SFT initialization as SafeDojo.
For model-based RL, we compare against WMPO~\cite{zhu2026wmpo} and WoVR~\cite{jiang2026wovr}, recent world-model-based policy optimizers for VLA without explicit safety constraints.

\noindent\textbf{Training Details.} For each suite, we use a fixed offline set of $1{,}500$ SafeLIBERO trajectories ($750$ collision and $750$ collision-free) to train world-model and safety components shared by its four tasks. All unsafe supervision comes from simulation; real-robot adaptation uses $50$ collision-free demonstrations per task and no unsafe robot data. We finetune the interactive world model from Wan2.2 for $5{,}000$ optimization steps with a learning rate of $1\times10^{-5}$.
The task reward classifier and the safety contact head are each trained for $20$ epochs with learning rates of $1\times10^{-4}$ and $3\times10^{-4}$, respectively.
For policy optimization, SafeDojo runs safety-constrained GRPO with group size $G=16$ and a learning rate of $2\times10^{-5}$, without collecting additional unsafe trajectories. Appendix~\ref{app:implementation_details} gives compute, checkpoint-granularity, and data-budget details.

\noindent\textbf{Metrics.} We evaluate methods using five metrics.
Task Success Rate (TSR) and Safe Success Rate (SSR) measure task completion without and with the collision-free requirement, respectively.
Collision Avoidance Rate (CAR) reports the percentage of episodes that complete without any collision.
Collision Step Count (CSC) reports the average number of in-contact simulation steps per episode, averaged over all evaluation episodes (including those with zero contact).
Execution Time Steps (ETS) measures efficiency as the average number of steps per episode.

\subsection{Simulation Results}

Table~\ref{tab:main_level1} and Table~\ref{tab:main_level2} present aggregate results on SafeLIBERO Level I and Level II, respectively. Complete per-task results are provided in Appendix Table~\ref{tab:per_task_level1} and Table~\ref{tab:per_task_level2}. 
SafeDojo achieves the best average task success and execution efficiency on both levels, and a clear safe-success improvement on Level I. 
We highlight several key findings below.

\noindent\textbf{SafeDojo improves safe success on Level I.}
SafeDojo obtains a $53.25$ average Safe Success Rate (SSR) on Level I, an $8.25$ percentage-point improvement over SafeVLA ($45.00$). On Level II, SafeDojo marginally leads SafeVLA ($49.62$ vs. $49.50$).
SSR is the most stringent metric, as it requires completing the instructed task without any obstacle collision. 
At the suite level, SafeDojo is strongest or competitive in SSR while maintaining strong task success. Held-out world-model and evaluator metrics are reported in Sec.~\ref{sec:component_validation}.

\noindent\textbf{SafeDojo improves execution efficiency.}
SafeDojo achieves the lowest average Execution Time Steps (ETS) of $269.41$ on Level I, ahead of the strongest baseline WMPO ($274.02$), indicating that safety-aware optimization does not make the policy overly conservative.
By reducing collision-prone actions and failed recovery behaviors, SafeDojo completes tasks with fewer unnecessary steps. In contrast, OpenVLA-OFT ($287.94$ ETS) and PPO ($285.67$ ETS) often waste steps on obstacle interference, collision recovery, or timeout failures.

\begin{table*}[t]
\centering
\caption{\textbf{Quantitative results on the SafeLIBERO Level I benchmark.} We compare SafeDojo against SFT, CBF, model-free RL, and model-based RL baselines. TSR: Task Success Rate; SSR: Safe Success Rate; ETS: Execution Time Steps. Best results are in \textbf{bold}, second best are \underline{underlined}.}
\label{tab:main_level1}
\resizebox{\textwidth}{!}{%
\begin{tabular}{llccccccccccccccc}
\toprule
& & \multicolumn{5}{c}{\textbf{TSR (\%) $\uparrow$}} & \multicolumn{5}{c}{\textbf{SSR (\%) $\uparrow$}} & \multicolumn{5}{c}{\textbf{ETS $\downarrow$}} \\
\cmidrule(lr){3-7} \cmidrule(lr){8-12} \cmidrule(lr){13-17}
\textbf{Type} & \textbf{Method} & Spa. & Goal & Obj. & Long & Avg. & Spa. & Goal & Obj. & Long & Avg. & Spa. & Goal & Obj. & Long & Avg. \\
\midrule
SFT & OpenVLA-OFT & 29.50 & 83.50 & 50.50 & \underline{45.00} & 52.12 & 22.50 & 73.50 & 26.00 & \textbf{34.00} & 39.00 & 263.49 & 170.35 & 252.46 & \underline{465.44} & 287.94 \\
CBF & AEGIS & 27.50 & 57.00 & 32.00 & 13.00 & 32.38 & 26.50 & 56.50 & 20.00 & 11.00 & 28.50 & 269.10 & 214.18 & 272.82 & 486.60 & 310.67 \\
\midrule
\multirow{2}{*}{\shortstack[l]{Model-free\\RL}}
 & PPO  & 35.00 & \underline{84.00} & 48.50 & \textbf{49.00} & 54.12 & 28.00 & 73.00 & 27.00 & \underline{31.50} & 39.88 & 253.67 & 170.84 & 256.92 & \textbf{461.24} & 285.67 \\
 & SafeVLA & 41.00 & 83.50 & 54.50 & 38.00 & 54.25 & 34.50 & \textbf{79.50} & 35.00 & 31.00 & \underline{45.00} & 236.70 & \textbf{166.97} & 249.98 & 525.10 & 294.69 \\
\midrule
\multirow{2}{*}{\shortstack[l]{Model-based\\RL}}
 & WMPO & \underline{54.00} & \textbf{85.50} & \underline{63.50} & 39.50 & \underline{60.62} & \underline{41.00} & 71.50 & \underline{37.50} & 29.50 & 44.88 & 217.74 & \underline{167.18} & \underline{238.82} & 472.32 & \underline{274.02} \\
 & WoVR & 53.50 & 77.00 & 29.00 & 43.00 & 50.62 & 40.00 & 65.50 & 14.00 & 31.00 & 37.62 & \underline{216.98} & 183.23 & 273.17 & 478.56 & 287.99 \\
\midrule
Ours & \textbf{SafeDojo} & \textbf{59.50} & 80.00 & \textbf{73.50} & \underline{45.00} & \textbf{64.50} & \textbf{51.50} & \underline{78.50} & \textbf{49.00} & \textbf{34.00} & \textbf{53.25} & \textbf{209.25} & 171.12 & \textbf{227.13} & 470.15 & \textbf{269.41} \\
\bottomrule
\end{tabular}%
}
\vspace{-0.5cm}
\end{table*}

\begin{table}[t]
\centering
\caption{\textbf{Quantitative results on SafeLIBERO Level II benchmark.} Methods are trained with Level I demonstrations only, without additional collection. TSR: Task Success Rate; SSR: Safe Success Rate; ETS: Execution Time Steps. Best results are in \textbf{bold}, second best are \underline{underlined}.}
\label{tab:main_level2}
\vskip 0.1in
\resizebox{\textwidth}{!}{%
\begin{tabular}{llccccccccccccccc}
\toprule
& & \multicolumn{5}{c}{\textbf{TSR (\%) $\uparrow$}} & \multicolumn{5}{c}{\textbf{SSR (\%) $\uparrow$}} & \multicolumn{5}{c}{\textbf{ETS $\downarrow$}} \\
\cmidrule(lr){3-7} \cmidrule(lr){8-12} \cmidrule(lr){13-17}
\textbf{Type} & \textbf{Method} & Spa. & Goal & Obj. & Long & Avg. & Spa. & Goal & Obj. & Long & Avg. & Spa. & Goal & Obj. & Long & Avg. \\
\midrule
SFT & OpenVLA-OFT & 57.00 & 62.00 & \underline{63.00} & 26.50 & 52.12 & 44.00 & 48.00 & \underline{57.50} & \underline{24.00} & 43.38 & 216.62 & 201.25 & 223.33 & 505.69 & 286.72 \\
CBF & AEGIS & 36.00 & 47.50 & 37.00 & 12.00 & 33.13 & 36.00 & 37.50 & 36.50 & 12.00 & 30.50 & 248.23 & 224.29 & 255.54 & \textbf{481.69} & 302.44 \\
\midrule
\multirow{2}{*}{\shortstack[l]{Model-free\\RL}}
 & PPO & 57.50 & \underline{64.50} & 62.00 & 24.50 & 52.12 & 47.00 & 48.50 & \underline{57.50} & 20.00 & 43.25 & 217.10 & 195.35 & \underline{222.30} & 507.28 & 285.51 \\
 & SafeVLA & 68.50 & \textbf{67.50} & \textbf{67.00} & 23.00 & \underline{56.50} & \textbf{61.00} & \underline{55.50} & \textbf{62.50} & 19.00 & \underline{49.50} & 194.68 & \textbf{189.60} & \textbf{221.31} & 553.87 & 289.87 \\
\midrule
\multirow{2}{*}{\shortstack[l]{Model-based\\RL}}
 & WMPO & \underline{72.00} & \underline{64.50} & 59.50 & \underline{27.00} & 55.75 & 58.50 & 48.00 & 54.00 & \underline{24.00} & 46.12 & \underline{188.63} & 192.27 & 228.23 & \underline{499.56} & \underline{277.17} \\
 & WoVR & 66.00 & 62.50 & 51.00 & 21.00 & 50.12 & 60.00 & 51.00 & 45.50 & 19.50 & 44.00 & 201.03 & 192.72 & 233.31 & 516.00 & 285.76 \\
\midrule
Ours & \textbf{SafeDojo} & \textbf{74.50} & 64.00 & 60.00 & \textbf{29.50} & \textbf{57.00} & \underline{60.50} & \textbf{56.00} & 55.50 & \textbf{26.50} & \textbf{49.62} & \textbf{187.50} & \underline{190.23} & 225.99 & 504.29 & \textbf{277.00} \\
\bottomrule
\end{tabular}%
}
\vspace{-0.5cm}
\end{table}

\noindent\textbf{SafeDojo generalizes beyond training safety scenarios.} To evaluate safety generalization, we train all methods using only Level I demonstrations and directly evaluate them on Level II, which introduces different safety scenarios by placing obstacles farther from the target but along the motion path. 
As shown in Table~\ref{tab:main_level2}, with complete per-task results in Appendix Table~\ref{tab:per_task_level2}, SafeDojo achieves the lowest average ETS ($277.00$; WMPO: $277.17$), the highest average TSR ($57.00$; SafeVLA: $56.50$), and the highest average SSR ($49.62$; SafeVLA: $49.50$) without additional Level II demonstrations. Although the SSR margin is small, these results show that the learned safety signals transfer to unseen obstacle configurations without retraining.

\subsection{Real-World Results}

\begin{figure}
    \centering
    \includegraphics[width=0.90\linewidth]{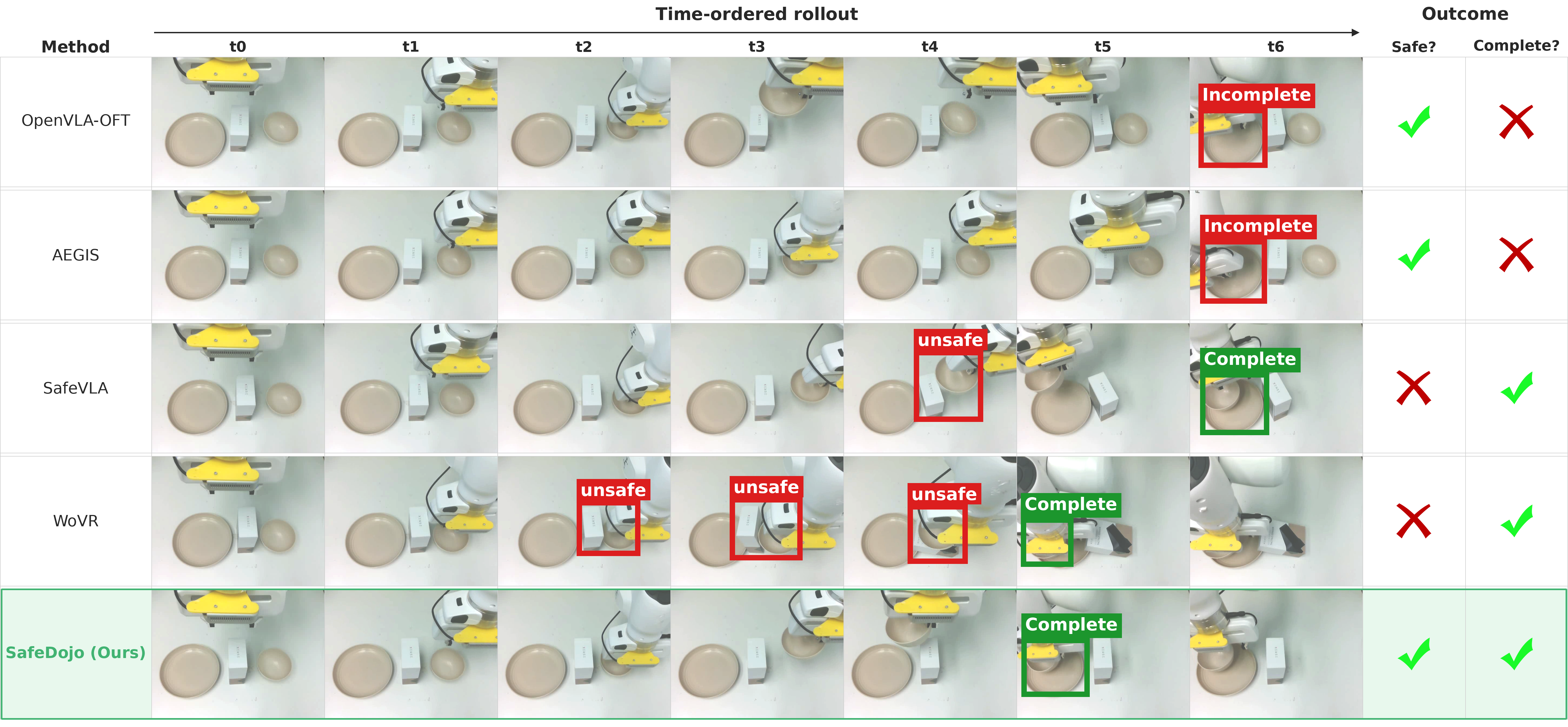}
    \caption{\textbf{Real-World Experiment Visualization.} SafeDojo completes the real-world task safely, while baselines either fail the task, violate safety, or succeed only with unsafe contacts.}
    \label{fig:real_world}
    \vspace{-0.3cm}
\end{figure}

To validate physical applicability, we deploy all methods on a Franka Emika Panda robot across five real-world manipulation tasks, including four single-arm tasks corresponding to the SafeLIBERO suites and one dual-arm coordination task. 
Figure~\ref{fig:real_world} visualizes representative executions that compare policy behaviors under obstacle-rich settings. 
Detailed real-world results are reported in Appendix Table~\ref{tab:real_world_results}, where SafeDojo achieves the best average TSR ($76.00$) and SSR ($70.00$) across five tasks.
SafeDojo consistently generates direct yet collision-aware trajectories, avoiding both obstacle contacts and unnecessary detours. 
In contrast, baselines either collide with obstacles, become overly conservative after safety filtering, or fail to recover from unsafe intermediate states. 
These quantitative results indicate that the reward--cost signals learned through imagined rollouts can transfer to physical robot execution, while the modest trial count warrants cautious interpretation.

\subsection{Ablation Study}

\begin{figure}[t]
    \centering
     \includegraphics[width=0.99\linewidth]{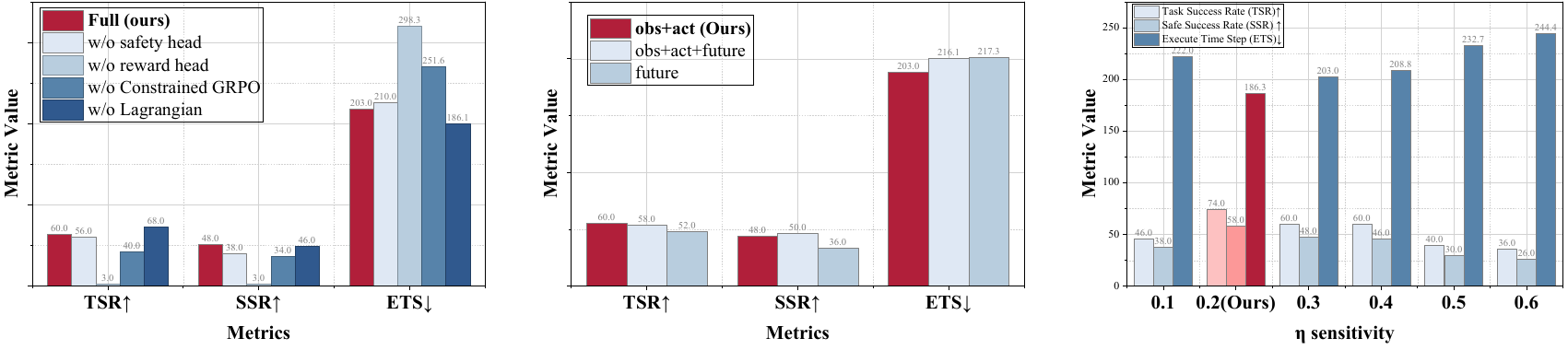}
    \caption{Ablation studies on SafeLIBERO Level I Spatial Task 0. Left: component ablation by removing each design choice independently. Center: input-source ablation comparing three configurations for reward--cost estimation. Right: sensitivity analysis of the initial safety cost weight $\eta$.}
    \label{fig:ablation}
    \vspace{-0.5cm}
\end{figure}

\noindent\textbf{Component ablation.}
We evaluate the contribution of each major component in SafeDojo as shown in Figure~\ref{fig:ablation}. For controlled component ablation, all variants share the same initialization $\eta_0=0.3$.
With this common initialization, the full model adaptively updates $\eta$ using Eq.~\ref{eq:lagrange_update} and achieves 60.0\% TSR, 48.0\% SSR, and 203.0 ETS. 
Removing the safety head reduces SSR to 38.0\%, confirming that safety-cost estimation is critical for safe-successful execution. 
Removing the task progress reward branch collapses TSR and SSR to 3.0\% and increases ETS to 298.3, indicating task-reward feedback is essential for goal-directed policy improvement. 
Replacing GRPO with unconstrained optimization decreases TSR and SSR to 40.0\% and 34.0\%, validating the importance of safety-constrained policy updates. 
Removing the adaptive Lagrangian multiplier increases TSR to 68.0\% but lowers SSR to 46.0\%, indicating a trade-off between task completion and safe success in this ablation.
Removing the safety constraint also reduces TSR because, in these obstacle-interference tasks, collision-prone shortcuts can obstruct motion, destabilize grasps, and displace target objects rather than merely trading safety for faster completion.

\noindent\textbf{Input source ablation.}
We investigate the effectiveness of different input sources for reward--cost estimation, as shown in Figure~\ref{fig:ablation}. For consistency with the component ablation, all input-source configurations are evaluated with the initial value of $\eta$ set to $0.3$. Here, the default observation--action input uses the current latent context $z^{\mathrm{ctx}}_{i,k}$ and proposed action chunk; ``future'' denotes additional predicted states within that chunk. Using future predictions alone yields a TSR of 52.0\%, an SSR of 36.0\%, and an ETS of 217.3, indicating that future states alone provide insufficient information for reliable task-progress and safety assessment. Adding these future predictions to the current imagined observation and proposed action raises SSR to 50.0\%, but decreases TSR to 58.0\% and increases ETS to 216.1, suggesting that accumulated prediction errors may introduce noisy evaluation signals. In contrast, the default observation--action input achieves the best overall trade-off, with a TSR of 60.0\%, an SSR of 48.0\%, and an ETS of 203.0. These results indicate that the current world-model state together with the proposed action chunk provides the most stable reward--cost cues, while additional future predictions offer only marginal safety gains and reduce overall stability.

\noindent\textbf{$\eta$ sensitivity.}
We study the effect of the initial safety weight $\eta$ on TSR, SSR, and ETS in Figure~\ref{fig:ablation}. Increasing $\eta$ from 0.1 to 0.2 improves TSR from 46.0\% to 74.0\% and SSR from 38.0\% to 58.0\%; increasing it to 0.6 instead lowers TSR/SSR to 36.0\%/26.0\%. Although Eq.~\ref{eq:lagrange_update} adapts $\eta$, only 20--100 policy updates are used, so initialization affects early rollout distributions and the path of the non-convex joint optimization. We use $\eta=0.2$. At the selected checkpoint, the mean top-$M$ predicted cost is $0.101$ against $d=0.2$, giving a surrogate constraint violation of $-0.099$; this does not establish satisfaction of the original CMDP constraint in deployment.

\subsection{Evaluation of Learned Components}
\label{sec:component_validation}
Beyond evaluating end-to-end policy performance, we directly assess the learned world model and evaluator heads. We measure video-prediction quality using FVD, PSNR, and SSIM, and evaluate the safety and success heads on autoregressively generated imagined inputs that match their use during policy optimization. All evaluations use data excluded from component training. Table~\ref{tab:component_validation} summarizes the results across the four SafeLIBERO suites.

\begin{table}[h]
\centering
\caption{Evaluation of the learned components across SafeLIBERO suites.}
\label{tab:component_validation}
\small
\setlength{\tabcolsep}{7pt}
\renewcommand{\arraystretch}{1.10}
\begin{tabular}{lcccc}
\toprule
\textbf{Metric} & \textbf{Spatial} & \textbf{Goal} & \textbf{Object} & \textbf{Long} \\
\midrule
FVD$\downarrow$ & 95.86 & 137.76 & 147.91 & 97.74 \\
PSNR$\uparrow$ & 24.63 & 22.08 & 23.45 & 21.94 \\
SSIM$\uparrow$ & 0.894 & 0.868 & 0.881 & 0.854 \\
Safety-head ROC-AUC$\uparrow$ & 96.98\% & 94.59\% & 92.67\% & 91.34\% \\
Success accuracy$\uparrow$ & 96.53\% & 97.14\% & 99.89\% & 98.37\% \\
Success ROC-AUC$\uparrow$ & 93.49\% & 98.62\% & 99.03\% & 95.05\% \\
\bottomrule
\end{tabular}
\end{table}

\vspace{-0.15cm}
\section{Limitations}
\label{sec:limitations}
Safety costs are inferred from visual world-model latents and actions with binary collision labels, which do not capture finer-grained risks such as contact force or obstacle proximity. Autoregressive world-model errors may also accumulate across action chunks, reducing the reliability of long-horizon reward and safety-cost estimates.

\vspace{-0.2cm}
\section{Conclusion}
\vspace{-0.1cm}
We present SafeDojo, a world model-based safe RL framework for VLA models that learns task-progress and safety signals from imagined trajectories without risky real-world exploration or hand-crafted safety functions. Experiments on SafeLIBERO and real-world deployment show strong safe-success performance on both levels and the best average TSR/SSR in our physical evaluation. These results support world-model-based safe RL as a promising mechanism for safer embodied intelligence.

\clearpage
\acknowledgments{This work was supported by the National Natural Science Foundation of China (62476011), the Beijing Natural Science Foundation (L252060).}

\bibliography{safedojo}

\begin{thebibliography}{44}
\providecommand{\natexlab}[1]{#1}
\providecommand{\url}[1]{\texttt{#1}}
\expandafter\ifx\csname urlstyle\endcsname\relax
  \providecommand{\doi}[1]{doi: #1}\else
  \providecommand{\doi}{doi: \begingroup \urlstyle{rm}\Url}\fi

\bibitem[Brunke et~al.(2022)Brunke, Greeff, Hall, Yuan, Zhou, Panerati, and
  Schoellig]{brunke2022safe}
L.~Brunke, M.~Greeff, A.~W. Hall, Z.~Yuan, S.~Zhou, J.~Panerati, and A.~P.
  Schoellig.
\newblock Safe learning in robotics: From learning-based control to safe
  reinforcement learning.
\newblock \emph{Annual Review of Control, Robotics, and Autonomous Systems},
  5:\penalty0 411--444, 2022.

\bibitem[Gu et~al.(2024)Gu, Yang, Du, Chen, Walter, Wang, and
  Knoll]{gu2024review}
S.~Gu, L.~Yang, Y.~Du, G.~Chen, F.~Walter, J.~Wang, and A.~Knoll.
\newblock A review of safe reinforcement learning: Methods, theories, and
  applications.
\newblock \emph{IEEE Transactions on Pattern Analysis and Machine
  Intelligence}, 46\penalty0 (12):\penalty0 11216--11235, 2024.

\bibitem[Khatib(1986)]{khatib1986real}
O.~Khatib.
\newblock Real-time obstacle avoidance for manipulators and mobile robots.
\newblock \emph{The International Journal of Robotics Research}, 5\penalty0
  (1):\penalty0 90--98, 1986.
\newblock \doi{10.1177/027836498600500106}.

\bibitem[Haddadin et~al.(2008)Haddadin, Albu-Sch{\"a}ffer, De~Luca, and
  Hirzinger]{haddadin2008collision}
S.~Haddadin, A.~Albu-Sch{\"a}ffer, A.~De~Luca, and G.~Hirzinger.
\newblock Collision detection and reaction: A contribution to safe physical
  human-robot interaction.
\newblock In \emph{2008 IEEE/RSJ International Conference on Intelligent Robots
  and Systems}, pages 3356--3363, 2008.
\newblock \doi{10.1109/IROS.2008.4650764}.

\bibitem[Haddadin et~al.(2009)Haddadin, Albu-Sch{\"a}ffer, and
  Hirzinger]{haddadin2009requirements}
S.~Haddadin, A.~Albu-Sch{\"a}ffer, and G.~Hirzinger.
\newblock Requirements for safe robots: Measurements, analysis and new
  insights.
\newblock \emph{The International Journal of Robotics Research}, 28\penalty0
  (11--12):\penalty0 1507--1527, 2009.
\newblock \doi{10.1177/0278364909343970}.

\bibitem[Haddadin et~al.(2017)Haddadin, De~Luca, and
  Albu-Sch{\"a}ffer]{haddadin2017robot}
S.~Haddadin, A.~De~Luca, and A.~Albu-Sch{\"a}ffer.
\newblock Robot collisions: A survey on detection, isolation, and
  identification.
\newblock \emph{IEEE Transactions on Robotics}, 33\penalty0 (6):\penalty0
  1292--1312, 2017.
\newblock \doi{10.1109/TRO.2017.2723903}.

\bibitem[Zanchettin et~al.(2016)Zanchettin, Ceriani, Rocco, Ding, and
  Matthias]{zanchettin2016safety}
A.~M. Zanchettin, N.~M. Ceriani, P.~Rocco, H.~Ding, and B.~Matthias.
\newblock Safety in human-robot collaborative manufacturing environments:
  Metrics and control.
\newblock \emph{IEEE Transactions on Automation Science and Engineering},
  13\penalty0 (2):\penalty0 882--893, 2016.
\newblock \doi{10.1109/TASE.2015.2412256}.

\bibitem[Lasota et~al.(2017)Lasota, Fong, and Shah]{lasota2017survey}
P.~A. Lasota, T.~Fong, and J.~A. Shah.
\newblock A survey of methods for safe human-robot interaction.
\newblock \emph{Foundations and Trends in Robotics}, 5\penalty0 (4):\penalty0
  261--349, 2017.
\newblock \doi{10.1561/2300000052}.

\bibitem[Ding et~al.(2022)Ding, Wang, Ren, Zheng, Chen, and He]{ding2022safety}
X.~Ding, H.~Wang, Y.~Ren, Y.~Zheng, C.~Chen, and J.~He.
\newblock Safety-critical optimal control for robotic manipulators in a
  cluttered environment.
\newblock \emph{arXiv preprint arXiv:2211.04944}, 2022.
\newblock \doi{10.48550/arXiv.2211.04944}.

\bibitem[Brohan et~al.(2023)Brohan, Brown, Carbajal, Chebotar, Chen,
  Choromanski, Ding, Driess, Dubey, Finn, et~al.]{brohan2023rt2}
A.~Brohan, N.~Brown, J.~Carbajal, Y.~Chebotar, X.~Chen, K.~Choromanski,
  T.~Ding, D.~Driess, A.~Dubey, C.~Finn, et~al.
\newblock Rt-2: Vision-language-action models transfer web knowledge to robotic
  control.
\newblock In \emph{Conference on Robot Learning}, 2023.

\bibitem[{Open X-Embodiment Collaboration} et~al.(2023)]{openx2023rtx}
{Open X-Embodiment Collaboration} et~al.
\newblock Open x-embodiment: Robotic learning datasets and rt-x models.
\newblock \emph{arXiv preprint arXiv:2310.08864}, 2023.

\bibitem[Kim et~al.(2024)Kim, Pertsch, Karamcheti, Xiao, Balakrishna, Nair,
  Rafailov, Foster, Lam, Sanketi, et~al.]{kim2024openvla}
M.~J. Kim, K.~Pertsch, S.~Karamcheti, T.~Xiao, A.~Balakrishna, S.~Nair,
  R.~Rafailov, E.~Foster, G.~Lam, P.~Sanketi, et~al.
\newblock Openvla: An open-source vision-language-action model.
\newblock \emph{arXiv preprint arXiv:2406.09246}, 2024.

\bibitem[Team et~al.(2024)]{octo2024}
O.~M. Team et~al.
\newblock Octo: An open-source generalist robot policy.
\newblock \emph{arXiv preprint arXiv:2405.12213}, 2024.

\bibitem[Black et~al.(2024)Black, Brown, Driess, Esmail, Equi, Finn, Fusai,
  Groom, Hausman, Ichter, et~al.]{black2024pi0}
K.~Black, N.~Brown, D.~Driess, A.~Esmail, M.~Equi, C.~Finn, N.~Fusai, L.~Groom,
  K.~Hausman, B.~Ichter, et~al.
\newblock $\pi_0$: A vision-language-action flow model for general robot
  control.
\newblock \emph{arXiv preprint arXiv:2410.24164}, 2024.

\bibitem[Ames et~al.(2019)Ames, Coogan, Egerstedt, Notomista, Sreenath, and
  Tabuada]{ames2019control}
A.~D. Ames, S.~Coogan, M.~Egerstedt, G.~Notomista, K.~Sreenath, and P.~Tabuada.
\newblock Control barrier functions: Theory and applications.
\newblock \emph{2019 18th European Control Conference (ECC)}, pages 3420--3431,
  2019.
\newblock \doi{10.23919/ECC.2019.8796030}.

\bibitem[Huang et~al.(2024)Huang, Ji, Xia, Zhang, and
  Yang]{huang2024safedreamer}
W.~Huang, J.~Ji, C.~Xia, B.~Zhang, and Y.~Yang.
\newblock {SafeDreamer}: Safe reinforcement learning with world models.
\newblock In \emph{International Conference on Learning Representations}, 2024.
\newblock URL \url{https://openreview.net/forum?id=tsE5HLYtYg}.

\bibitem[Zhang et~al.(2025)Zhang, Zhang, Ji, Lei, Dai, Chen, and
  Yang]{zhang2025safevla}
B.~Zhang, Y.~Zhang, J.~Ji, Y.~Lei, J.~Dai, Y.~Chen, and Y.~Yang.
\newblock {SafeVLA}: Towards safety alignment of vision-language-action model
  via constrained learning.
\newblock In \emph{Thirty-ninth Conference on Neural Information Processing
  Systems}, 2025.
\newblock URL \url{https://openreview.net/forum?id=dt940loCBT}.
\newblock Spotlight.

\bibitem[Hu et~al.(2025)Hu, Liu, Liu, Cen, Meng, and He]{hu2025vlsa}
S.~Hu, Z.~Liu, S.~Liu, J.~Cen, Z.~Meng, and X.~He.
\newblock {VLSA}: Vision-language-action models with plug-and-play safety
  constraint layer.
\newblock \emph{arXiv preprint arXiv:2512.11891}, 2025.
\newblock URL \url{https://arxiv.org/abs/2512.11891}.

\bibitem[Cao et~al.(2025)Cao, Xin, Wu, He, Yan, Tan, and Wang]{cao2025fosp}
C.~Cao, Y.~Xin, S.~Wu, L.~He, Z.~Yan, J.~Tan, and X.~Wang.
\newblock {FOSP}: Fine-tuning offline safe policy through world models.
\newblock In \emph{International Conference on Learning Representations}, 2025.
\newblock URL \url{https://openreview.net/forum?id=dbuFJg7eaw}.

\bibitem[Yu et~al.(2026)Yu, Zhou, Huang, Khadiv, and Yang]{yu2026safenight}
D.~Yu, Q.~Zhou, B.~Huang, M.~Khadiv, and Z.~Yang.
\newblock Safe-night {VLA}: Seeing the unseen via thermal-perceptive
  vision-language-action models for safety-critical manipulation.
\newblock \emph{arXiv preprint arXiv:2603.05754}, 2026.
\newblock URL \url{https://arxiv.org/abs/2603.05754}.

\bibitem[Son et~al.(2026)Son, Ko, Choi, and Lim]{son2026thermoact}
Y.-C. Son, D.-K. Ko, Y.-J. Choi, and S.-C. Lim.
\newblock {ThermoAct}: Thermal-aware vision-language-action models for robotic
  perception and decision-making.
\newblock \emph{IEEE Robotics and Automation Letters}, 11\penalty0
  (5):\penalty0 6106--6113, 2026.
\newblock \doi{10.1109/LRA.2026.3678130}.

\bibitem[Zhai et~al.(2026)Zhai, Ou, Yu, Hao, and Liu]{zhai2026cofreevla}
X.~Zhai, B.~Ou, Q.~Yu, C.~Hao, and Y.~Liu.
\newblock {CoFreeVLA}: Collision-free dual-arm manipulation via
  vision-language-action model and risk estimation, 2026.
\newblock URL \url{https://arxiv.org/abs/2601.21712}.

\bibitem[Gu et~al.(2025)Gu, Ju, Sun, Gilitschenski, Nishimura, Itkina, and
  Shkurti]{gu2025safe}
Q.~Gu, Y.~Ju, S.~Sun, I.~Gilitschenski, H.~Nishimura, M.~Itkina, and
  F.~Shkurti.
\newblock {SAFE}: Multitask failure detection for vision-language-action
  models.
\newblock \emph{arXiv preprint arXiv:2506.09937}, 2025.
\newblock \doi{10.48550/arXiv.2506.09937}.
\newblock URL \url{https://arxiv.org/abs/2506.09937}.

\bibitem[Zhang et~al.(2024)Zhang, Zheng, Chen, Jang, Li, Han, Wang, Ding, Fox,
  and Yao]{zhang2024grape}
Z.~Zhang, K.~Zheng, Z.~Chen, J.~Jang, Y.~Li, S.~Han, C.~Wang, M.~Ding, D.~Fox,
  and H.~Yao.
\newblock {GRAPE}: Generalizing robot policy via preference alignment.
\newblock \emph{arXiv preprint arXiv:2411.19309}, 2024.
\newblock \doi{10.48550/arXiv.2411.19309}.
\newblock URL \url{https://arxiv.org/abs/2411.19309}.

\bibitem[Altman(1999)]{altman1999constrained}
E.~Altman.
\newblock \emph{Constrained Markov Decision Processes}.
\newblock Chapman and Hall/CRC, 1999.

\bibitem[Achiam et~al.(2017)Achiam, Held, Tamar, and
  Abbeel]{achiam2017constrained}
J.~Achiam, D.~Held, A.~Tamar, and P.~Abbeel.
\newblock Constrained policy optimization.
\newblock In \emph{Proceedings of the 34th International Conference on Machine
  Learning}, volume~70 of \emph{Proceedings of Machine Learning Research},
  pages 22--31. PMLR, 2017.
\newblock URL \url{https://proceedings.mlr.press/v70/achiam17a.html}.

\bibitem[Ray et~al.(2019)Ray, Achiam, and Amodei]{ray2019benchmarking}
A.~Ray, J.~Achiam, and D.~Amodei.
\newblock Benchmarking safe exploration in deep reinforcement learning.
\newblock OpenAI technical report, 2019.
\newblock URL
  \url{https://openai.com/index/benchmarking-safe-exploration-in-deep-reinforcement-learning/}.

\bibitem[Tessler et~al.(2019)Tessler, Mankowitz, and Mannor]{tessler2019rcpo}
C.~Tessler, D.~J. Mankowitz, and S.~Mannor.
\newblock Reward constrained policy optimization.
\newblock In \emph{International Conference on Learning Representations}, 2019.
\newblock URL \url{https://openreview.net/forum?id=SkfrvsA9FX}.

\bibitem[Zhang et~al.(2020)Zhang, Vuong, and Ross]{zhang2020focops}
Y.~Zhang, Q.~Vuong, and K.~Ross.
\newblock First order constrained optimization in policy space.
\newblock In \emph{Advances in Neural Information Processing Systems},
  volume~33, 2020.
\newblock URL
  \url{https://proceedings.neurips.cc/paper/2020/hash/af5d5ef24881f3c3049a7b9bfe74d58b-Abstract.html}.

\bibitem[Stooke et~al.(2020)Stooke, Achiam, and Abbeel]{stooke2020responsive}
A.~Stooke, J.~Achiam, and P.~Abbeel.
\newblock Responsive safety in reinforcement learning by {PID} lagrangian
  methods.
\newblock In \emph{Proceedings of the 37th International Conference on Machine
  Learning}, volume 119 of \emph{Proceedings of Machine Learning Research},
  pages 9133--9143. PMLR, 2020.
\newblock URL \url{https://proceedings.mlr.press/v119/stooke20a.html}.

\bibitem[Thomas et~al.(2021)Thomas, Luo, and Ma]{thomas2021safe}
G.~Thomas, Y.~Luo, and T.~Ma.
\newblock Safe reinforcement learning by imagining the near future.
\newblock In \emph{Advances in Neural Information Processing Systems},
  volume~34, 2021.
\newblock URL \url{https://openreview.net/forum?id=vIDBSGl3vzl}.

\bibitem[Hogewind et~al.(2023)Hogewind, Sim{\~a}o, Kachman, and
  Jansen]{hogewind2023safeslac}
Y.~Hogewind, T.~D. Sim{\~a}o, T.~Kachman, and N.~Jansen.
\newblock Safe reinforcement learning from pixels using a stochastic latent
  representation.
\newblock In \emph{International Conference on Learning Representations}, 2023.
\newblock URL \url{https://openreview.net/forum?id=b39dQt_uffW}.

\bibitem[Nakamura et~al.(2025)Nakamura, Peters, and Bajcsy]{nakamura2025latent}
K.~Nakamura, L.~Peters, and A.~Bajcsy.
\newblock Generalizing safety beyond collision-avoidance via latent-space
  reachability analysis.
\newblock In \emph{Proceedings of Robotics: Science and Systems}, Los Angeles,
  CA, USA, June 2025.
\newblock \doi{10.15607/RSS.2025.XXI.113}.
\newblock URL \url{https://www.roboticsproceedings.org/rss21/p113.html}.

\bibitem[Seo et~al.(2025)Seo, Nakamura, and Bajcsy]{seo2025unisafe}
J.~Seo, K.~Nakamura, and A.~Bajcsy.
\newblock Uncertainty-aware latent safety filters for avoiding
  out-of-distribution failures.
\newblock In \emph{Proceedings of The 9th Conference on Robot Learning}, volume
  305 of \emph{Proceedings of Machine Learning Research}, pages 4442--4472.
  PMLR, 2025.
\newblock URL \url{https://proceedings.mlr.press/v305/seo25a.html}.

\bibitem[Zhu et~al.(2026)Zhu, Yan, Hong, Shou, Ma, and Guo]{zhu2026wmpo}
F.~Zhu, Z.~Yan, Z.~Hong, Q.~Shou, X.~Ma, and S.~Guo.
\newblock {WMPO}: World model-based policy optimization for
  vision-language-action models.
\newblock In \emph{International Conference on Learning Representations}, 2026.
\newblock URL \url{https://openreview.net/forum?id=qE2FyvRvuF}.

\bibitem[Jiang et~al.(2026)Jiang, Zhou, Jiang, Huang, Wei, Chen, Zhou, Guo,
  Lin, Zhang, Wang, Li, Yu, and Zhao]{jiang2026wovr}
Z.~Jiang, S.~Zhou, Y.~Jiang, Z.~Huang, M.~Wei, Y.~Chen, T.~Zhou, Z.~Guo,
  H.~Lin, Q.~Zhang, Y.~Wang, H.~Li, C.~Yu, and D.~Zhao.
\newblock {WoVR}: World models as reliable simulators for post-training {VLA}
  policies with {RL}.
\newblock \emph{arXiv preprint arXiv:2602.13977}, 2026.
\newblock URL \url{https://arxiv.org/abs/2602.13977}.

\bibitem[Xiao et~al.(2025)Xiao, Yang, Chang, Chen, Xiong, Xu, Zheng, and
  Zhang]{xiao2025worldenv}
J.~Xiao, Y.~Yang, X.~Chang, R.~Chen, F.~Xiong, M.~Xu, W.-S. Zheng, and
  Q.~Zhang.
\newblock {World-Env}: Leveraging world model as a virtual environment for
  {VLA} post-training.
\newblock \emph{arXiv preprint arXiv:2509.24948}, 2025.
\newblock URL \url{https://arxiv.org/abs/2509.24948}.

\bibitem[Li et~al.(2025)Li, Ding, Suo, Wang, Ge, Zang, Yu, Sun, Zhang, Wang,
  and Su]{li2025vlarft}
H.~Li, P.~Ding, R.~Suo, Y.~Wang, Z.~Ge, D.~Zang, K.~Yu, M.~Sun, H.~Zhang,
  D.~Wang, and W.~Su.
\newblock {VLA-RFT}: Vision-language-action reinforcement fine-tuning with
  verified rewards in world simulators.
\newblock \emph{arXiv preprint arXiv:2510.00406}, 2025.
\newblock \doi{10.48550/arXiv.2510.00406}.
\newblock URL \url{https://arxiv.org/abs/2510.00406}.

\bibitem[Liu et~al.(2026)Liu, Bai, Ci, Ma, and Shou]{liu2026worldvlaloop}
X.~Liu, Z.~Bai, H.~Ci, K.~Y. Ma, and M.~Z. Shou.
\newblock {World-VLA-Loop}: Closed-loop learning of video world model and {VLA}
  policy.
\newblock \emph{arXiv preprint arXiv:2602.06508}, 2026.
\newblock \doi{10.48550/arXiv.2602.06508}.
\newblock URL \url{https://arxiv.org/abs/2602.06508}.

\bibitem[Wan et~al.(2025)Wan, Wang, Ai, Wen, Mao, Xie, Chen, Yu, Zhao, Yang,
  Zeng, Wang, Zhang, Zhou, Wang, Chen, Zhu, Zhao, Yan, Huang, Feng, Zhang, Li,
  Wu, Chu, Feng, Zhang, Sun, Fang, Wang, Gui, Weng, Shen, Lin, Wang, Wang,
  Zhou, Wang, Shen, Yu, Shi, Huang, Xu, Kou, Lv, Li, Liu, Wang, Zhang, Huang,
  Li, Wu, Liu, Pan, Zheng, Hong, Shi, Feng, Jiang, Han, Wu, and Liu]{wan2025}
T.~Wan, A.~Wang, B.~Ai, B.~Wen, C.~Mao, C.-W. Xie, D.~Chen, F.~Yu, H.~Zhao,
  J.~Yang, J.~Zeng, J.~Wang, J.~Zhang, J.~Zhou, J.~Wang, J.~Chen, K.~Zhu,
  K.~Zhao, K.~Yan, L.~Huang, M.~Feng, N.~Zhang, P.~Li, P.~Wu, R.~Chu, R.~Feng,
  S.~Zhang, S.~Sun, T.~Fang, T.~Wang, T.~Gui, T.~Weng, T.~Shen, W.~Lin,
  W.~Wang, W.~Wang, W.~Zhou, W.~Wang, W.~Shen, W.~Yu, X.~Shi, X.~Huang, X.~Xu,
  Y.~Kou, Y.~Lv, Y.~Li, Y.~Liu, Y.~Wang, Y.~Zhang, Y.~Huang, Y.~Li, Y.~Wu,
  Y.~Liu, Y.~Pan, Y.~Zheng, Y.~Hong, Y.~Shi, Y.~Feng, Z.~Jiang, Z.~Han, Z.-F.
  Wu, and Z.~Liu.
\newblock Wan: Open and advanced large-scale video generative models.
\newblock \emph{arXiv preprint arXiv:2503.20314}, 2025.

\bibitem[Girgis et~al.(2026)Girgis, de~Schaetzen, Rowe, Robitaille, Pal, and
  Paull]{girgis2026constrainedgrpo}
R.~Girgis, R.~de~Schaetzen, L.~Rowe, A.~Robitaille, C.~Pal, and L.~Paull.
\newblock Constrained group relative policy optimization, 2026.
\newblock URL \url{https://arxiv.org/abs/2602.05863}.

\bibitem[Liu et~al.(2023)Liu, Zhu, Gao, Feng, Liu, Zhu, and
  Stone]{liu2023libero}
B.~Liu, Y.~Zhu, C.~Gao, Y.~Feng, Q.~Liu, Y.~Zhu, and P.~Stone.
\newblock {LIBERO}: Benchmarking knowledge transfer for lifelong robot
  learning.
\newblock In \emph{Thirty-seventh Conference on Neural Information Processing
  Systems Datasets and Benchmarks Track}, 2023.
\newblock URL \url{https://openreview.net/forum?id=xzEtNSuDJk}.

\bibitem[Zhu et~al.(2020)Zhu, Wong, Mandlekar, Mart{\'i}n-Mart{\'i}n, Joshi,
  Lin, Maddukuri, Nasiriany, and Zhu]{zhu2020robosuite}
Y.~Zhu, J.~Wong, A.~Mandlekar, R.~Mart{\'i}n-Mart{\'i}n, A.~Joshi, K.~Lin,
  A.~Maddukuri, S.~Nasiriany, and Y.~Zhu.
\newblock {robosuite}: A modular simulation framework and benchmark for robot
  learning.
\newblock \emph{arXiv preprint arXiv:2009.12293}, 2020.
\newblock URL \url{https://arxiv.org/abs/2009.12293}.

\bibitem[Kim et~al.(2025)Kim, Finn, and Liang]{kim2025fine}
M.~J. Kim, C.~Finn, and P.~Liang.
\newblock Fine-tuning vision-language-action models: Optimizing speed and
  success.
\newblock In \emph{Proceedings of Robotics: Science and Systems}, Los Angeles,
  CA, USA, June 2025.
\newblock \doi{10.15607/RSS.2025.XXI.017}.
\newblock URL \url{https://www.roboticsproceedings.org/rss21/p017.html}.

\end{thebibliography}

\newpage
\appendix

\section{Overview of Appendices}
\begin{itemize}
    \item Appendix~\ref{app:related_work}: Detailed Related Work
    \item Appendix~\ref{app:broader_impact}: Broader Impact
    \item Appendix~\ref{app:level1_results}: Per-Task Results on SafeLIBERO Level I
    \item Appendix~\ref{app:level2_results}: Per-Task Results on SafeLIBERO Level II
    \item Appendix~\ref{app:task_descriptions}: Task Descriptions
    \item Appendix~\ref{app:real_world_results}: Real-World Experiment Details
    \item Appendix~\ref{app:implementation_details}: Implementation Details
    \item Appendix~\ref{app:reward_cost_heads}: Reward and Safety Cost Heads
\end{itemize}

\section{Detailed Related Work}
\label{app:related_work}

\noindent\textbf{Safety for VLA models.}
Existing safety mechanisms for vision-language-action (VLA) models mainly fall
into perception augmentation, inference-time intervention, and policy-level
alignment. Perception-oriented methods extend RGB VLAs with additional sensing
modalities, such as Safe-Night VLA~\cite{yu2026safenight} and
ThermoAct~\cite{son2026thermoact}, improving robustness under safety-critical
conditions but requiring specialized sensors or modality-specific pipelines.
Inference-time methods monitor or constrain actions after generation:
AEGIS/VLSA~\cite{hu2025vlsa} applies a CBF-based safety layer over geometric
constraints, CoFreeVLA~\cite{zhai2026cofreevla} gates unsafe dual-arm motions,
and SAFE~\cite{gu2025safe} predicts execution failures from VLA features.
These approaches improve deployment-time safety, but do not optimize the VLA
with predicted long-horizon safety costs. Policy-level methods such as
SafeVLA~\cite{zhang2025safevla} and GRAPE~\cite{zhang2024grape} incorporate
safety into alignment or optimization, yet do not leverage an action-conditioned
video world model to anticipate future visual safety costs during optimization.
In contrast, SafeDojo studies constrained VLA optimization with
action-conditioned future safety costs predicted by a video world model.

\noindent\textbf{Model-Based Safe RL.}
Safe RL is commonly formulated as a CMDP that maximizes reward under cost
constraints~\cite{altman1999constrained}. Model-free constrained policy
optimization methods~\cite{achiam2017constrained, ray2019benchmarking,
tessler2019rcpo, zhang2020focops, stooke2020responsive} suffer from sample
inefficiency and unsafe exploration, while model-based methods evaluate
candidate behaviors through imagined rollouts or enforce constraints with
latent safety filters~\cite{thomas2021safe, huang2024safedreamer,
hogewind2023safeslac, cao2025fosp, nakamura2025latent, seo2025unisafe}.
However, these methods are designed for generic control and usually rely on
predefined scalar costs, rather than task-conditioned visual safety costs for
language-conditioned manipulation. Meanwhile, world-model-based VLA
post-training~\cite{zhu2026wmpo, jiang2026wovr, xiao2025worldenv,
li2025vlarft, liu2026worldvlaloop} optimizes task return in imagination but
does not impose explicit CMDP-style safety constraints. SafeDojo bridges these
directions by using an action-conditioned video world model to estimate both
progress from predicted frames and collision risk from latent representations,
enabling CMDP-style safe VLA optimization in imagination.

\begin{table}[t]
\centering
\caption{Comparison of safe reinforcement learning families along two key axes for VLA deployment. 
(i) \emph{Anticipatory}, whether the method avoids constraint violations before they occur in the real world; 
(ii) \emph{Learned safety}, whether the safety signal is learned from data rather than hand-crafted. 
}
\label{tab:safe_rl_taxonomy}
\small
\setlength{\tabcolsep}{6pt}
\renewcommand{\arraystretch}{1.2}
\resizebox{\textwidth}{!}{%
\begin{tabular}{llcc}
\toprule
\textbf{Family} & \textbf{Representative methods} & \textbf{Anticipatory} & \textbf{Learned safety} \\
\midrule
Lagrangian / CMDP        & CPO, PPO-Lag, FOCOPS, SafeVLA          & \xmarkc & \xmarkc \\
Control-theoretic        & CBF-QP, AEGIS, Safe-Night VLA          & \cmarkc & \xmarkc \\
Model-based (state-based)    & SMBPO, FOSP               & \cmarkc & \xmarkc \\
Gaussian process         & Safe-GP, SAMBA                         & \xmarkc & \xmarkc \\
\midrule
\rowcolor{gray!15}
\textbf{Model-based (video)} & \textbf{SafeDojo (ours)} & \cmarkc & \cmarkc \\
\bottomrule
\end{tabular}
}
\end{table}

\begin{table*}[h]
\centering
\caption{Per-task results on SafeLIBERO Level I. We compare SafeDojo against SFT, CBF, model-free RL, and model-based RL baselines. All methods use OpenVLA-OFT as the base policy. CAR ($\uparrow$), TSR ($\uparrow$), and SSR ($\uparrow$) are reported as percentages, while ETS ($\downarrow$) and CSC ($\downarrow$) are reported in steps. Best results are in \textbf{bold}, second best are \underline{underlined}. See Table~\ref{tab:task_descriptions} for task full descriptions.}
\label{tab:per_task_level1}
\scriptsize
\setlength{\tabcolsep}{2.0pt}
\renewcommand{\arraystretch}{1.08}
\resizebox{\textwidth}{!}{%
\begin{tabular}{@{}llcccccccccccccccc@{}}
\toprule
\multirow{2}{*}{\textbf{Method}} & \multirow{2}{*}{\textbf{Metric}} & \multicolumn{4}{c}{\textbf{Spatial}} & \multicolumn{4}{c}{\textbf{Goal}} & \multicolumn{4}{c}{\textbf{Object}} & \multicolumn{4}{c}{\textbf{Long}} \\
\cmidrule(lr){3-6} \cmidrule(lr){7-10} \cmidrule(lr){11-14} \cmidrule(lr){15-18}
& & \textbf{Task 0} & \textbf{Task 1} & \textbf{Task 2} & \textbf{Task 3} & \textbf{Task 0} & \textbf{Task 1} & \textbf{Task 2} & \textbf{Task 3} & \textbf{Task 0} & \textbf{Task 1} & \textbf{Task 2} & \textbf{Task 3} & \textbf{Task 0} & \textbf{Task 1} & \textbf{Task 2} & \textbf{Task 3} \\
\midrule
\multirow{5}{*}{OpenVLA-OFT} & CAR$\uparrow$ & 42.00 & 52.00 & 80.00 & 24.00 & 88.00 & 64.00 & 86.00 & \underline{96.00} & 64.00 & 28.00 & 62.00 & 34.00 & 38.00 & 88.00 & 90.00 & \underline{86.00} \\
 & TSR$\uparrow$ & 20.00 & 30.00 & 54.00 & 14.00 & 92.00 & \underline{82.00} & 92.00 & \underline{68.00} & 30.00 & 40.00 & 68.00 & 64.00 & 60.00 & 20.00 & \underline{62.00} & \underline{38.00} \\
 & SSR$\uparrow$ & 10.00 & 24.00 & 48.00 & 8.00 & 84.00 & 62.00 & \underline{80.00} & \textbf{68.00} & 24.00 & 20.00 & 38.00 & 22.00 & 24.00 & 16.00 & \underline{60.00} & \underline{36.00} \\
 & ETS$\downarrow$ & 266.40 & 260.74 & 237.68 & 289.14 & 126.98 & 153.26 & \underline{154.26} & 246.90 & 268.18 & 273.68 & 255.30 & 212.68 & 461.02 & 532.36 & \textbf{405.34} & \underline{463.06} \\
 & CSC$\downarrow$ & 36.12 & 51.92 & 3.20 & 21.66 & 3.32 & 13.36 & 4.20 & \underline{2.10} & 23.84 & 34.44 & 1.06 & 24.10 & 34.16 & 4.70 & 16.62 & 16.72 \\
\midrule
\multirow{5}{*}{AEGIS} & CAR$\uparrow$ & \textbf{98.00} & \underline{62.00} & \textbf{90.00} & \textbf{66.00} & \textbf{98.00} & \textbf{100.00} & \textbf{98.00} & \textbf{98.00} & 40.00 & \textbf{52.00} & \textbf{98.00} & \underline{46.00} & \textbf{88.00} & 72.00 & \textbf{98.00} & \textbf{100.00} \\
 & TSR$\uparrow$ & 14.00 & 24.00 & 46.00 & 26.00 & 72.00 & \underline{82.00} & 38.00 & 36.00 & 2.00 & \textbf{68.00} & 10.00 & 48.00 & 48.00 & 0.00 & 4.00 & 0.00 \\
 & SSR$\uparrow$ & 14.00 & 24.00 & 46.00 & 22.00 & 72.00 & \textbf{82.00} & 36.00 & 36.00 & 0.00 & \textbf{38.00} & 10.00 & 32.00 & \textbf{40.00} & 0.00 & 4.00 & 0.00 \\
 & ETS$\downarrow$ & 290.10 & 277.82 & 239.92 & 268.54 & 180.70 & 147.24 & 256.08 & 272.68 & 298.92 & 255.74 & 295.92 & 240.68 & \underline{455.66} & \textbf{500.00} & 490.74 & 500.00 \\
 & CSC$\downarrow$ & \textbf{0.20} & 42.88 & \textbf{1.66} & 47.70 & \textbf{0.02} & \textbf{0.00} & \textbf{0.02} & 4.64 & 38.90 & \textbf{2.68} & \textbf{0.08} & 28.74 & \textbf{1.24} & 4.34 & \textbf{0.24} & \textbf{0.00} \\
\midrule
\multirow{5}{*}{PPO} & CAR$\uparrow$ & 46.00 & 46.00 & 76.00 & 20.00 & 90.00 & 64.00 & 84.00 & 94.00 & 60.00 & 24.00 & 60.00 & 40.00 & 24.00 & 84.00 & 86.00 & 84.00 \\
 & TSR$\uparrow$ & 38.00 & 30.00 & 60.00 & 12.00 & \underline{96.00} & 80.00 & \underline{94.00} & 66.00 & 18.00 & 46.00 & 66.00 & 64.00 & \textbf{72.00} & \textbf{30.00} & 54.00 & \textbf{40.00} \\
 & SSR$\uparrow$ & 32.00 & 28.00 & 48.00 & 4.00 & 88.00 & 62.00 & 78.00 & 64.00 & 18.00 & 22.00 & 36.00 & 32.00 & 14.00 & \textbf{24.00} & 50.00 & \textbf{38.00} \\
 & ETS$\downarrow$ & 246.20 & 256.62 & 222.86 & 289.00 & \underline{122.90} & 154.04 & 155.96 & 250.48 & 281.86 & 272.54 & 259.52 & 213.74 & \textbf{443.68} & \underline{522.88} & 419.58 & \textbf{458.82} \\
 & CSC$\downarrow$ & 31.00 & 44.44 & 3.86 & 22.80 & 2.96 & 12.56 & 4.46 & 3.08 & 27.78 & 38.06 & 0.92 & 30.38 & 26.98 & 8.82 & 22.90 & 29.60 \\
\midrule
\multirow{5}{*}{SafeVLA} & CAR$\uparrow$ & 64.00 & 58.00 & \underline{86.00} & 30.00 & \underline{96.00} & 76.00 & \underline{94.00} & 94.00 & \underline{66.00} & 32.00 & \underline{70.00} & \textbf{54.00} & \underline{46.00} & \textbf{94.00} & \underline{94.00} & \underline{86.00} \\
 & TSR$\uparrow$ & 46.00 & 48.00 & 54.00 & 16.00 & \textbf{100.00} & 80.00 & \textbf{96.00} & 58.00 & 24.00 & 50.00 & 66.00 & \underline{78.00} & 56.00 & 20.00 & 36.00 & \textbf{40.00} \\
 & SSR$\uparrow$ & \underline{38.00} & 42.00 & 52.00 & 6.00 & \textbf{96.00} & 74.00 & \textbf{92.00} & 56.00 & 22.00 & 26.00 & 44.00 & \textbf{48.00} & \underline{32.00} & 20.00 & 36.00 & \underline{36.00} \\
 & ETS$\downarrow$ & 217.92 & 220.46 & 224.62 & 283.80 & \textbf{117.84} & 153.56 & \textbf{143.40} & 253.08 & 276.08 & 270.30 & 260.28 & \underline{193.28} & 513.52 & 583.60 & 512.14 & 491.16 \\
 & CSC$\downarrow$ & 36.58 & 54.94 & 6.90 & 18.56 & 1.92 & 7.96 & \underline{0.98} & 2.44 & 22.38 & 32.24 & \underline{0.82} & \underline{14.16} & 32.78 & 5.58 & \underline{5.28} & \underline{14.70} \\
\midrule
\multirow{5}{*}{WMPO} & CAR$\uparrow$ & 48.00 & 54.00 & 78.00 & 38.00 & 86.00 & 62.00 & 80.00 & 92.00 & 58.00 & 30.00 & 68.00 & 42.00 & 36.00 & 84.00 & 92.00 & 82.00 \\
 & TSR$\uparrow$ & 56.00 & \textbf{58.00} & \textbf{70.00} & 32.00 & 94.00 & \textbf{88.00} & 90.00 & \textbf{70.00} & \underline{42.00} & \underline{64.00} & \underline{76.00} & 72.00 & 52.00 & \underline{22.00} & 50.00 & 34.00 \\
 & SSR$\uparrow$ & \underline{38.00} & \underline{48.00} & \underline{60.00} & 18.00 & 82.00 & 62.00 & 76.00 & \underline{66.00} & \underline{36.00} & 30.00 & \underline{50.00} & 34.00 & 20.00 & 16.00 & 50.00 & 32.00 \\
 & ETS$\downarrow$ & 209.96 & \textbf{207.86} & \textbf{199.18} & 253.96 & 125.68 & \textbf{144.28} & 156.24 & \textbf{242.50} & \underline{255.20} & \textbf{250.64} & \underline{247.10} & 202.32 & 458.14 & 534.32 & 425.80 & 471.02 \\
 & CSC$\downarrow$ & 30.00 & \underline{34.64} & 3.30 & \underline{11.54} & 3.32 & 14.14 & 8.08 & \textbf{0.98} & \underline{22.26} & 30.66 & 0.96 & 17.36 & \underline{20.24} & \underline{3.56} & 19.94 & 25.22 \\
\midrule
\multirow{5}{*}{WoVR} & CAR$\uparrow$ & 48.00 & 50.00 & 78.00 & 48.00 & 92.00 & 56.00 & 76.00 & 92.00 & 24.00 & 26.00 & 46.00 & 22.00 & 38.00 & \underline{90.00} & 84.00 & 82.00 \\
 & TSR$\uparrow$ & \underline{58.00} & 44.00 & \textbf{70.00} & \textbf{42.00} & 86.00 & 80.00 & 76.00 & 66.00 & 14.00 & 18.00 & 52.00 & 32.00 & \underline{64.00} & \underline{22.00} & 46.00 & \textbf{40.00} \\
 & SSR$\uparrow$ & 36.00 & 36.00 & \underline{60.00} & \underline{28.00} & 84.00 & 50.00 & 64.00 & 64.00 & 8.00 & 6.00 & 24.00 & 18.00 & 22.00 & \underline{22.00} & 46.00 & 34.00 \\
 & ETS$\downarrow$ & \underline{196.56} & 227.12 & 202.90 & \underline{241.32} & 143.60 & 165.66 & 177.28 & \underline{246.38} & 286.62 & 286.76 & 263.56 & 255.74 & 460.18 & 535.84 & 451.14 & 467.10 \\
 & CSC$\downarrow$ & 26.30 & 47.92 & 4.46 & 15.52 & \underline{0.94} & 13.10 & 13.44 & 2.30 & 37.66 & 33.48 & 5.24 & 58.56 & 22.86 & 4.66 & 14.14 & 23.14 \\
\midrule
\multirow{5}{*}{\textbf{SafeDojo}} & CAR$\uparrow$ & \underline{72.00} & \textbf{64.00} & 84.00 & \underline{58.00} & 94.00 & \underline{80.00} & 84.00 & \underline{96.00} & \textbf{72.00} & \underline{34.00} & \underline{70.00} & \textbf{54.00} & 36.00 & \underline{90.00} & 90.00 & \underline{86.00} \\
 & TSR$\uparrow$ & \textbf{74.00} & \underline{56.00} & \underline{68.00} & \underline{40.00} & 94.00 & \underline{82.00} & 80.00 & 64.00 & \textbf{60.00} & 62.00 & \textbf{90.00} & \textbf{82.00} & 60.00 & \underline{22.00} & \textbf{64.00} & 34.00 \\
 & SSR$\uparrow$ & \textbf{58.00} & \textbf{50.00} & \textbf{64.00} & \textbf{34.00} & \underline{94.00} & \underline{78.00} & 78.00 & 64.00 & \textbf{56.00} & \underline{34.00} & \textbf{60.00} & \underline{46.00} & 24.00 & 18.00 & \textbf{62.00} & 32.00 \\
 & ETS$\downarrow$ & \textbf{186.26} & \underline{209.33} & \underline{200.53} & \textbf{240.88} & 125.32 & \underline{145.52} & 164.67 & 248.98 & \textbf{232.44} & \underline{254.76} & \textbf{238.90} & \textbf{182.44} & 466.34 & 534.76 & \underline{409.78} & 469.74 \\
 & CSC$\downarrow$ & \underline{8.72} & \textbf{26.88} & \underline{2.20} & \textbf{10.86} & 1.53 & \underline{5.45} & 4.13 & 2.12 & \textbf{15.02} & \underline{15.38} & 0.94 & \textbf{8.62} & 36.32 & \textbf{3.28} & 13.08 & 15.28 \\
\bottomrule
\end{tabular}%
}
\end{table*}

\section{Broader Impact}
\label{app:broader_impact}

SafeDojo advances the deployment of vision-language-action 
models on physical robots by enabling safer interaction with 
real-world environments. This contributes to the broader 
goal of making embodied AI systems more reliable and 
trustworthy for applications such as household assistance, 
industrial automation, and human-robot collaboration. By 
reducing the risk of physical violations during policy 
learning, our framework lowers the barrier for deploying 
learning-based robotic systems beyond controlled laboratory 
settings.

We also acknowledge potential risks. Safety mechanisms 
learned from data may exhibit unexpected failure modes when 
deployed in scenarios that differ substantially from training 
distributions, and overreliance on learned safety signals 
without external safeguards could lead to false confidence 
in autonomous systems. We therefore advocate for combining 
SafeDojo with complementary safety measures, including 
runtime monitoring and human oversight, in real-world 
deployment. Furthermore, as with any advance in robotic 
autonomy, the technologies enabled by safer VLA deployment 
should be developed with consideration for ethical, 
economic, and societal implications.

\begin{table*}[h]
\centering
\caption{Per-task results on SafeLIBERO Level II (generalization). All methods are trained with Level I demonstrations only, without any additional data collection. CAR ($\uparrow$), TSR ($\uparrow$), and SSR ($\uparrow$) are reported as percentages, while ETS ($\downarrow$) and CSC ($\downarrow$) are reported in steps. Best results are in \textbf{bold}, second best are \underline{underlined}. See Table~\ref{tab:task_descriptions} for task full descriptions.}
\label{tab:per_task_level2}
\scriptsize
\setlength{\tabcolsep}{2.0pt}
\renewcommand{\arraystretch}{1.08}
\resizebox{\textwidth}{!}{%
\begin{tabular}{@{}llcccccccccccccccc@{}}
\toprule
\multirow{2}{*}{\textbf{Method}} & \multirow{2}{*}{\textbf{Metric}} & \multicolumn{4}{c}{\textbf{Spatial}} & \multicolumn{4}{c}{\textbf{Goal}} & \multicolumn{4}{c}{\textbf{Object}} & \multicolumn{4}{c}{\textbf{Long}} \\
\cmidrule(lr){3-6} \cmidrule(lr){7-10} \cmidrule(lr){11-14} \cmidrule(lr){15-18}
& & \textbf{Task 0} & \textbf{Task 1} & \textbf{Task 2} & \textbf{Task 3} & \textbf{Task 0} & \textbf{Task 1} & \textbf{Task 2} & \textbf{Task 3} & \textbf{Task 0} & \textbf{Task 1} & \textbf{Task 2} & \textbf{Task 3} & \textbf{Task 0} & \textbf{Task 1} & \textbf{Task 2} & \textbf{Task 3} \\
\midrule
\multirow{5}{*}{OpenVLA-OFT} & CAR$\uparrow$ & 78.00 & 68.00 & 64.00 & 84.00 & 54.00 & 84.00 & 62.00 & 38.00 & 86.00 & 80.00 & \underline{96.00} & 70.00 & 70.00 & 74.00 & 42.00 & 80.00 \\
 & TSR$\uparrow$ & 74.00 & 42.00 & 64.00 & 48.00 & 44.00 & 80.00 & \textbf{58.00} & 66.00 & \underline{70.00} & \textbf{74.00} & \underline{66.00} & 42.00 & \underline{30.00} & 36.00 & \textbf{24.00} & 16.00 \\
 & SSR$\uparrow$ & 62.00 & 26.00 & 40.00 & 48.00 & 40.00 & 76.00 & 44.00 & 32.00 & \underline{68.00} & 60.00 & \underline{64.00} & 38.00 & \underline{30.00} & 36.00 & \textbf{14.00} & 16.00 \\
 & ETS$\downarrow$ & 177.82 & 240.74 & 216.38 & 231.56 & 235.46 & 155.16 & 213.94 & 200.44 & 201.96 & \textbf{202.98} & \underline{239.30} & 249.08 & \textbf{498.20} & 486.46 & 512.38 & 525.74 \\
 & CSC$\downarrow$ & 1.92 & 4.98 & 6.98 & 6.88 & 33.26 & 8.70 & 6.68 & 10.48 & 11.36 & 2.40 & 2.86 & 14.88 & \underline{14.90} & 8.74 & 50.50 & 27.30 \\
\midrule
\multirow{5}{*}{AEGIS} & CAR$\uparrow$ & \underline{92.00} & \textbf{92.00} & \textbf{100.00} & \textbf{92.00} & \textbf{62.00} & \textbf{98.00} & 50.00 & \textbf{64.00} & \textbf{94.00} & \textbf{98.00} & \textbf{100.00} & \textbf{96.00} & \textbf{98.00} & \textbf{98.00} & \textbf{82.00} & \textbf{100.00} \\
 & TSR$\uparrow$ & 22.00 & 28.00 & 38.00 & 56.00 & 32.00 & \underline{90.00} & 36.00 & 32.00 & 30.00 & 62.00 & 8.00 & 48.00 & 2.00 & 2.00 & 4.00 & \textbf{40.00} \\
 & SSR$\uparrow$ & 22.00 & 28.00 & 38.00 & 56.00 & 14.00 & \underline{90.00} & 22.00 & 24.00 & 28.00 & \underline{62.00} & 8.00 & \underline{48.00} & 2.00 & 2.00 & 4.00 & \textbf{40.00} \\
 & ETS$\downarrow$ & 269.20 & 252.88 & 251.34 & 219.50 & 250.92 & \underline{143.98} & 252.54 & 249.72 & 264.50 & 230.78 & 294.18 & \textbf{232.70} & \underline{499.02} & 497.38 & \textbf{491.76} & \textbf{438.58} \\
 & CSC$\downarrow$ & 0.98 & \underline{2.54} & \textbf{0.00} & 6.02 & \textbf{16.42} & \underline{0.78} & 15.36 & 10.88 & \textbf{3.68} & \textbf{0.50} & \textbf{0.00} & \textbf{0.94} & \textbf{0.46} & \textbf{1.06} & \textbf{14.62} & \textbf{0.00} \\
\midrule
\multirow{5}{*}{PPO} & CAR$\uparrow$ & 74.00 & 66.00 & 66.00 & \underline{90.00} & 56.00 & 82.00 & 62.00 & 34.00 & 84.00 & 80.00 & \underline{96.00} & 72.00 & 68.00 & 72.00 & 38.00 & \underline{86.00} \\
 & TSR$\uparrow$ & 76.00 & 30.00 & 62.00 & 62.00 & \underline{52.00} & 82.00 & \underline{56.00} & 68.00 & 68.00 & \underline{72.00} & 56.00 & 52.00 & 26.00 & 38.00 & 18.00 & 16.00 \\
 & SSR$\uparrow$ & 62.00 & 18.00 & 46.00 & 62.00 & \underline{44.00} & 74.00 & 44.00 & 32.00 & 66.00 & \underline{62.00} & 56.00 & 46.00 & 24.00 & 36.00 & 4.00 & 16.00 \\
 & ETS$\downarrow$ & 181.40 & 255.80 & 214.50 & 216.70 & 218.10 & 152.00 & 217.10 & 194.20 & \underline{201.00} & 208.20 & 243.30 & \underline{236.70} & 500.10 & 477.70 & 519.60 & 531.70 \\
 & CSC$\downarrow$ & 3.30 & 5.74 & 8.76 & 2.88 & 42.42 & 8.06 & 6.66 & 16.24 & 11.98 & 2.80 & 0.72 & 10.78 & 29.02 & 13.00 & 59.22 & 23.90 \\
\midrule
\multirow{5}{*}{SafeVLA} & CAR$\uparrow$ & \underline{92.00} & 80.00 & \underline{78.00} & \textbf{92.00} & 58.00 & \textbf{98.00} & \underline{66.00} & 42.00 & \underline{90.00} & \underline{86.00} & \textbf{100.00} & \underline{80.00} & 66.00 & 72.00 & 36.00 & 78.00 \\
 & TSR$\uparrow$ & \underline{82.00} & 46.00 & 74.00 & \underline{72.00} & \textbf{54.00} & \textbf{94.00} & 52.00 & \underline{70.00} & 66.00 & \textbf{74.00} & \textbf{70.00} & \textbf{58.00} & 26.00 & 30.00 & \underline{22.00} & 14.00 \\
 & SSR$\uparrow$ & \underline{78.00} & \underline{38.00} & \underline{56.00} & \underline{72.00} & \underline{44.00} & \textbf{94.00} & 44.00 & \underline{40.00} & 66.00 & \textbf{64.00} & \textbf{70.00} & \textbf{50.00} & 26.00 & 30.00 & 6.00 & 14.00 \\
 & ETS$\downarrow$ & 158.52 & 227.84 & 199.56 & \underline{192.78} & 217.04 & \textbf{131.58} & 221.96 & \underline{187.80} & 202.22 & 206.62 & \textbf{238.70} & 237.70 & 552.40 & 531.54 & 553.70 & 577.82 \\
 & CSC$\downarrow$ & \textbf{0.26} & 3.72 & 4.24 & 2.38 & 36.96 & \textbf{0.58} & 6.86 & 11.62 & \underline{7.30} & \underline{0.52} & \textbf{0.00} & \underline{6.32} & 16.42 & 14.20 & 71.34 & 31.34 \\
\midrule
\multirow{5}{*}{WMPO} & CAR$\uparrow$ & 74.00 & 76.00 & 62.00 & 88.00 & 52.00 & \underline{88.00} & 58.00 & 36.00 & 78.00 & 76.00 & 92.00 & 72.00 & 66.00 & 62.00 & 30.00 & 82.00 \\
 & TSR$\uparrow$ & \textbf{86.00} & \underline{60.00} & 72.00 & 70.00 & 46.00 & 86.00 & 50.00 & \textbf{76.00} & 68.00 & 66.00 & 50.00 & \underline{54.00} & 22.00 & \underline{40.00} & \textbf{24.00} & \underline{22.00} \\
 & SSR$\uparrow$ & 68.00 & \textbf{48.00} & 48.00 & 70.00 & 38.00 & 82.00 & 38.00 & 34.00 & \underline{68.00} & 50.00 & 48.00 & \textbf{50.00} & 22.00 & \underline{40.00} & \underline{12.00} & \underline{22.00} \\
 & ETS$\downarrow$ & \textbf{149.34} & \underline{206.38} & 198.08 & 200.70 & 223.58 & 147.62 & 221.42 & \textbf{176.44} & 203.86 & 210.28 & 257.64 & 241.14 & 507.50 & \underline{465.98} & \underline{501.58} & \underline{523.20} \\
 & CSC$\downarrow$ & 2.12 & 2.84 & 2.60 & \underline{1.80} & 32.48 & 9.86 & 4.78 & \textbf{6.40} & 19.14 & 2.34 & 1.24 & 9.54 & 23.02 & 22.62 & 73.50 & 24.02 \\
\midrule
\multirow{5}{*}{WoVR} & CAR$\uparrow$ & \textbf{96.00} & \underline{84.00} & \underline{78.00} & \textbf{92.00} & 54.00 & 84.00 & 58.00 & 44.00 & 78.00 & 78.00 & 92.00 & 72.00 & 66.00 & \underline{82.00} & \underline{50.00} & 80.00 \\
 & TSR$\uparrow$ & \textbf{86.00} & 28.00 & \textbf{78.00} & \underline{72.00} & \underline{52.00} & 80.00 & \underline{56.00} & 62.00 & 60.00 & 70.00 & 48.00 & 26.00 & 20.00 & 36.00 & 14.00 & 14.00 \\
 & SSR$\uparrow$ & \textbf{84.00} & 24.00 & \textbf{60.00} & \underline{72.00} & 40.00 & 78.00 & \underline{48.00} & 38.00 & 58.00 & 54.00 & 44.00 & 26.00 & 20.00 & 36.00 & 8.00 & 14.00 \\
 & ETS$\downarrow$ & \underline{151.48} & 264.20 & \textbf{193.42} & 195.04 & \underline{216.18} & 152.28 & \underline{204.22} & 198.18 & 216.24 & \underline{206.40} & 248.98 & 261.64 & 511.32 & 495.00 & 526.24 & 531.42 \\
 & CSC$\downarrow$ & \underline{0.68} & 3.46 & \underline{1.32} & \textbf{1.30} & 24.36 & 11.18 & \underline{4.50} & 8.82 & 13.46 & 0.98 & 0.62 & 12.44 & 19.66 & 9.58 & \underline{37.58} & \underline{10.58} \\
\midrule
\multirow{5}{*}{\textbf{SafeDojo}} & CAR$\uparrow$ & 80.00 & 80.00 & 66.00 & \textbf{92.00} & \underline{60.00} & \underline{88.00} & \textbf{68.00} & \underline{50.00} & 82.00 & 82.00 & \underline{96.00} & 72.00 & \underline{74.00} & 80.00 & 32.00 & 78.00 \\
 & TSR$\uparrow$ & 80.00 & \textbf{62.00} & \underline{76.00} & \textbf{80.00} & \textbf{54.00} & 82.00 & \textbf{58.00} & 62.00 & \textbf{72.00} & 68.00 & 48.00 & 52.00 & \textbf{32.00} & \textbf{52.00} & 14.00 & 20.00 \\
 & SSR$\uparrow$ & 68.00 & \textbf{48.00} & 46.00 & \textbf{80.00} & \textbf{46.00} & 80.00 & \textbf{54.00} & \textbf{44.00} & \textbf{70.00} & 56.00 & 48.00 & \underline{48.00} & \textbf{32.00} & \textbf{52.00} & 6.00 & 16.00 \\
 & ETS$\downarrow$ & 164.28 & \textbf{205.44} & \underline{196.24} & \textbf{184.06} & \textbf{214.34} & 148.23 & \textbf{202.12} & 196.24 & \textbf{199.74} & 207.42 & 251.64 & 245.16 & 503.62 & \textbf{458.88} & 527.36 & 527.30 \\
 & CSC$\downarrow$ & 2.74 & \textbf{1.60} & 4.44 & 2.30 & \underline{20.13} & 2.24 & \textbf{4.02} & \underline{6.66} & 12.12 & 3.00 & \underline{0.36} & 14.24 & 21.56 & \underline{4.32} & 60.58 & 19.68 \\
\bottomrule
\end{tabular}%
}
\end{table*}

\section{Per-Task Results on SafeLIBERO Level I}
\label{app:level1_results}

Table~\ref{tab:per_task_level1} breaks down the Level I evaluation by task and
suite. Level I corresponds to the training obstacle setting, where obstacles
are placed close to task-relevant objects. The per-task metrics separate task
completion from safe completion: a large gap between TSR and SSR indicates that
the policy can often finish the instruction but still contacts the obstacle.

\section{Per-Task Results on SafeLIBERO Level II}
\label{app:level2_results}

Table~\ref{tab:per_task_level2} reports zero-shot generalization to Level II
obstacle layouts. All checkpoints are trained only with Level I demonstrations
and are evaluated without collecting additional Level II data. These results
therefore measure whether each method transfers its task and safety behavior to
unseen obstacle placements along the execution path.

\section{Task Descriptions}
\label{app:task_descriptions}

Table~\ref{tab:task_descriptions} maps the task indices used in the per-task
tables to their abbreviated names and full instructions. The four suites test complementary
manipulation capabilities: Spatial tasks vary the initial object location, Goal
tasks vary the desired placement target, Object tasks vary the manipulated
object identity, and Long tasks require completing two sequential placements.

\begin{table}[htbp]
\centering
\caption{Full task descriptions for SafeLIBERO.}
\label{tab:task_descriptions}
\small
\renewcommand{\arraystretch}{1.2}
\resizebox{\columnwidth}{!}{%
\begin{tabular}{llll}
\toprule
\textbf{Suite} & \textbf{Task ID} & \textbf{Abbreviation} & \textbf{Full Description} \\
\midrule
\multirow{4}{*}{Spatial}
 & Task 0 & Bowl: ramekin$\to$plate     & Pick up the black bowl between the ramekin and the plate and place it on the plate \\
 & Task 1 & Bowl: on ramekin$\to$plate  & Pick up the black bowl on the ramekin and place it on the plate \\
 & Task 2 & Bowl: stove$\to$plate       & Pick up the bowl on the stove and place it on the plate \\
 & Task 3 & Bowl: cabinet$\to$plate     & Pick up the bowl on the cabinet and place it on the plate \\
\midrule
\multirow{5}{*}{Goal}
 & Task 0 & Bowl$\to$plate              & Put the bowl on the plate \\
 & Task 1 & Bowl$\to$cabinet            & Put the bowl on top of the cabinet \\
 & Task 2 & Bowl$\to$stove              & Put the bowl on the stove \\
 & Task 3 (Level I) & Drawer$\to$bowl inside      & Open the top drawer and put the bowl inside \\
 & Task 3 (Level II) & Cream cheese$\to$bowl       & Put the cream cheese in the bowl \\
\midrule
\multirow{4}{*}{Object}
 & Task 0 & OJ$\to$basket               & Pick up the orange juice and place it in the basket \\
 & Task 1 & Pudding$\to$basket          & Pick up the chocolate pudding and place it in the basket \\
 & Task 2 & Milk$\to$basket             & Pick up the milk and place it in the basket \\
 & Task 3 & BBQ sauce$\to$basket        & Pick up the bbq sauce and place it in the basket \\
\midrule
\multirow{4}{*}{Long}
 & Task 0 & Soup+cheese$\to$basket      & Put both the alphabet soup and the cream cheese box in the basket \\
 & Task 1 & Soup+sauce$\to$basket       & Put both the alphabet soup and the tomato sauce in the basket \\
 & Task 2 & White mug+yellow mug        & Put the white mug on the left plate and put the yellow and white mug on the right plate \\
 & Task 3 & White mug+pudding           & Put the white mug on the plate and put the chocolate pudding to the right of the plate \\
\bottomrule
\end{tabular}%
}
\end{table}

\section{Real-World Experiment Details}
\label{app:real_world_results}

We evaluate real-world transfer on five tabletop manipulation tasks with
Franka Emika Panda robots. Each task is designed to require collision-aware
transport rather than only endpoint reaching. During execution, we place a
box-shaped obstacle on the nominal motion path between the manipulated object
and the goal region, so the robot must complete the instruction while avoiding
contact with the obstacle. Table~\ref{tab:real_world_task_descriptions}
summarizes the task set.

\begin{table}[h]
\centering
\caption{Real-world task descriptions. All tasks include a box obstacle placed along the execution path.}
\label{tab:real_world_task_descriptions}
\small
\renewcommand{\arraystretch}{1.2}
\begin{tabularx}{\textwidth}{@{}>{\raggedright\arraybackslash}p{0.21\textwidth}>{\raggedright\arraybackslash}X@{}}
\toprule
\textbf{Task} & \textbf{Description} \\
\midrule
Bowl-to-plate & Pick up the bowl and place it on the plate. \\
Apple selection & Select the apple from multiple objects, grasp it, and place it on the plate. \\
Bread-to-toaster & Pick up the bread slice and place it into the toaster. \\
Block stacking & Place the blue block on the plate, then stack the green block on top of the blue block. \\
Dual-arm stacking & The right arm places the green block on the plate, then the left arm stacks the blue block on top of the green block. \\
\bottomrule
\end{tabularx}
\end{table}

These tasks cover five representative physical behaviors: direct object
transport, object selection among distractors, insertion into a constrained
goal region, sequential manipulation with a stacking constraint, and dual-arm
coordination under obstacle interference. A trial is
considered task-successful if the final object configuration satisfies the
language instruction, and safe-successful only if the task is completed without
contacting the obstacle. This mirrors the SafeLIBERO evaluation protocol while
testing whether the learned reward-cost signals transfer to physical execution.
For each method-task pair, we conduct $10$ real-world trials and compute TSR
and SSR from these trials.

\begin{table*}[t]
\centering
\caption{Real-world quantitative results across five Franka manipulation tasks. We include the same methods as Table~\ref{tab:main_level1}. Each method-task pair is evaluated over $10$ trials. TSR and SSR are reported as percentages, where SSR requires both task completion and no contact with the box obstacle.}
\label{tab:real_world_results}
\small
\setlength{\tabcolsep}{4pt}
\renewcommand{\arraystretch}{1.15}
\resizebox{\textwidth}{!}{%
\begin{tabular}{llcccccccccccc}
\toprule
\multirow{2}{*}{\textbf{Type}} & \multirow{2}{*}{\textbf{Method}}
& \multicolumn{2}{c}{\textbf{Bowl-to-plate}}
& \multicolumn{2}{c}{\textbf{Apple selection}}
& \multicolumn{2}{c}{\textbf{Bread-to-toaster}}
& \multicolumn{2}{c}{\textbf{Block stacking}}
& \multicolumn{2}{c}{\textbf{Dual-arm stacking}}
& \multicolumn{2}{c}{\textbf{Average}} \\
\cmidrule(lr){3-4} \cmidrule(lr){5-6} \cmidrule(lr){7-8} \cmidrule(lr){9-10} \cmidrule(lr){11-12} \cmidrule(lr){13-14}
& & TSR & SSR & TSR & SSR & TSR & SSR & TSR & SSR & TSR & SSR & TSR & SSR \\
\midrule
SFT & OpenVLA-OFT & 50.00 & 20.00 & 40.00 & 20.00 & 10.00 & 10.00 & 0.00 & 0.00 & 0.00 & 0.00 & 20.00 & 10.00 \\
CBF & AEGIS & 30.00 & 30.00 & 40.00 & 40.00 & 20.00 & 10.00 & 0.00 & 0.00 & 0.00 & 0.00 & 18.00 & 16.00 \\
\multirow{2}{*}{\shortstack[l]{Model-free\\RL}}
& PPO & 60.00 & 20.00 & 50.00 & 30.00 & 30.00 & 20.00 & 30.00 & 0.00 & 10.00 & 0.00 & 36.00 & 14.00 \\
& SafeVLA & 60.00 & 50.00 & 60.00 & 60.00 & 40.00 & 30.00 & 30.00 & 30.00 & 10.00 & 10.00 & 40.00 & 36.00 \\
\multirow{2}{*}{\shortstack[l]{Model-based\\RL}}
& WMPO & 70.00 & 30.00 & 70.00 & 40.00 & 60.00 & 10.00 & 30.00 & 10.00 & 30.00 & 0.00 & 52.00 & 18.00 \\
& WoVR & 90.00 & 40.00 & 90.00 & 40.00 & 60.00 & 20.00 & 50.00 & 10.00 & 40.00 & 0.00 & 66.00 & 22.00 \\
\rowcolor{green!10}
Ours & \textbf{SafeDojo} & \textbf{90.00} & \textbf{90.00} & \textbf{100.00} & \textbf{90.00} & \textbf{70.00} & \textbf{70.00} & \textbf{60.00} & \textbf{50.00} & \textbf{60.00} & \textbf{50.00} & \textbf{76.00} & \textbf{70.00} \\
\bottomrule
\end{tabular}%
}
\end{table*}

\begin{figure*}[t]
    \centering
    \includegraphics[width=\textwidth]{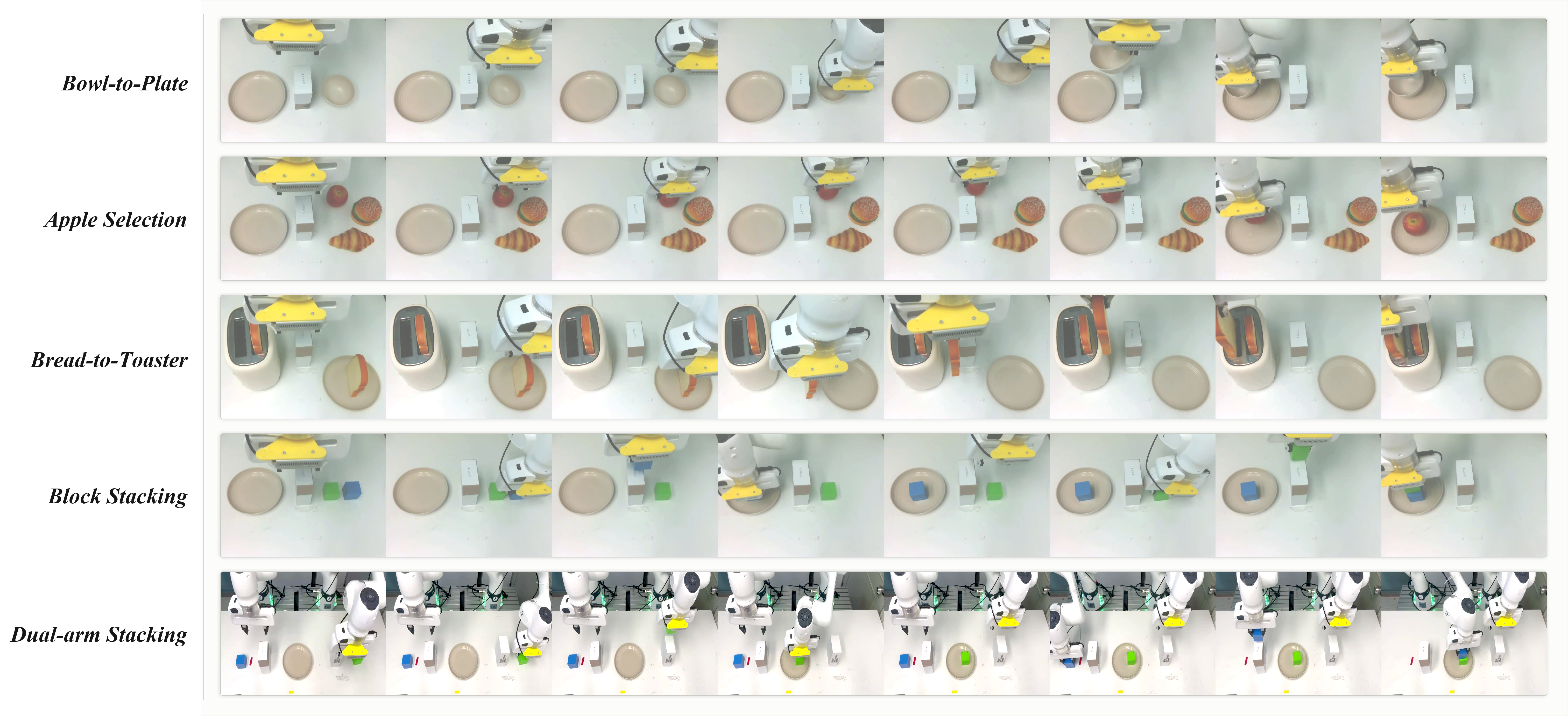}
    \caption{\textbf{Representative SafeDojo Real-World Demos.}
    Representative snapshots from SafeDojo executions on five
    real-world tasks.}
    \label{fig:real_world_appendix_demos}
\end{figure*}

\section{Implementation Details}
\label{app:implementation_details}

\noindent\textbf{Observation and action format.}
All simulation experiments use $256\times256$ RGB observations from the
SafeLIBERO camera setup. The VLA policy follows the OpenVLA-OFT action
interface and predicts 7-dimensional actions, consisting of a 6-DoF
end-effector delta pose and a gripper command. Actions are normalized with the
corresponding LIBERO/SafeLIBERO action statistics. Policy actions are executed
in chunks of $H=8$ low-level control steps, matching the world-model rollout
interface used by SafeDojo.

\noindent\textbf{World-model training.}
The interactive world model is initialized from Wan2.2 and fine-tuned on
the fixed per-suite set of $1{,}500$ SafeLIBERO rollout trajectories, evenly
split between collision and collision-free episodes, converted into the Wan/RLinf video-action
format. Each training sample contains $13$ video frames at $256\times256$
resolution, with $5$ conditioning frames and action conditioning supplied as an
extra input. We train only the DiT component while keeping the VAE fixed. The
world model is optimized with learning rate $1\times10^{-5}$ for $5{,}000$
steps, using static-video augmentation with probability $0.05$ to improve
robustness to self-generated rollout frames. During imagined rollouts, the
world model uses the current imagined observation, the latent context, the
language instruction, and the next 8-step action chunk to predict eight aligned
future frames and latent states.

\noindent\textbf{Policy optimization.}
SafeDojo starts from the same OpenVLA-OFT supervised fine-tuning checkpoint as
all baselines. Policy optimization is implemented with FSDP and AdamW. Unless
otherwise stated, we use learning rate $2\times10^{-5}$, Adam coefficients
$(0.9,0.999)$, Adam epsilon $1\times10^{-5}$, weight decay $0.01$, gradient
clipping at $1.0$, micro-batch size $32$, and global batch size $8192$. The
GRPO objective uses token-level log probabilities, token-mean loss
aggregation, advantage normalization, clipping ratios $0.2$ and $0.28$, and no
KL penalty in the reported runs. The safety-constrained variant aggregates the
largest $16$ predicted per-step costs when computing the trajectory-level safety
signal.
For checkpoint granularity, WoVR is trained at the task-suite level, using one
checkpoint for the corresponding SafeLIBERO suite. In contrast, WMPO and
SafeDojo use task-specific training, where each task has its own optimized
checkpoint.

\noindent\textbf{Unsafe-data budget.}
All collision supervision is collected offline in SafeLIBERO simulation. The
same $750$ unsafe trajectories per suite train the suite-shared world model and
safety head, and imagined policy optimization collects no additional unsafe
experience. As a reference point, one final SafeVLA Spatial checkpoint
accumulated $4{,}282$ unsafe trajectories, or $5.71\times$ this per-suite
budget. Real-robot adaptation mixes the simulation data with $50$
collision-free demonstrations per task; no unsafe real-robot trajectories are
collected.

\noindent\textbf{Rollout and evaluation protocol.}
For simulated training rollouts, SafeDojo uses world-model environments with
maximum rollout lengths of $240$ steps for Spatial, Goal, and Object suites and
$512$ steps for Long-horizon tasks before policy-level overrides.
Evaluation in the real SafeLIBERO simulator uses the corresponding SafeLIBERO environments
with $256\times256$ camera observations. We evaluate each method on both Level I
and Level II obstacle settings and report TSR, SSR, CAR, CSC, and ETS as
defined in Sec.~4. Real-world evaluation follows the protocol in
Appendix~\ref{app:real_world_results}, with $10$ trials for each method-task
pair.

\noindent\textbf{Compute and model sizes.}
The stack contains a 14B-parameter world model, a 7B VLA, an 86.2M task-success
classifier, and a 1.0M safety head. World-model fine-tuning uses 8 A800 GPUs;
policy optimization uses 32 A800-80GB GPUs across four nodes; both heads use a
single GPU for 20 epochs. All policy baselines use the same number of policy
updates.

\section{Reward and Safety Cost Heads}
\label{app:reward_cost_heads}

\noindent\textbf{Task reward head.}
The task branch $f_{\mathrm{task}}$ is implemented as a ResNet-style binary
success classifier compatible with the reward model interface used in the
world-model environment. The classifier takes a $256\times256$ RGB frame,
normalizes pixel values to $[-1,1]$, and outputs a scalar success logit. The
model is initialized from the LIBERO/WoVR reward-model checkpoint and fine-tuned
on SafeLIBERO rollout frames. Each frame is labeled with the recorded
\texttt{success\_per\_step} signal from the rollout trajectory, so positive
labels indicate that the task has reached a successful state at that step. We
train the classifier for $20$ epochs with batch size $256$ and learning rate
$1\times10^{-4}$ using \texttt{BCEWithLogitsLoss}. Because successful frames are
sparse, the loss uses a positive-class weight computed from the negative-to-
positive ratio in the training split. At inference time, the sigmoid output is
used as the dense task-progress reward for imagined frames.

\noindent\textbf{Safety cost head.}
The safety branch $f_{\mathrm{safe}}$ predicts obstacle-contact probability for each of
the $H=8$ action steps in a proposed action chunk. Its input is the current Wan
latent context $z^{\mathrm{ctx}}_{i,k}$, available before rolling out that chunk,
with 48 channels, two context latent slots, and $16\times16$ spatial resolution. The action-conditioned
head follows the ACSSP-style cost-head design: a compact latent encoder produces
a scene representation, an action-conditioning module fuses the proposed action
chunk, and an MLP head outputs per-step contact logits. In our default
imagined-observation--action mode, the model uses scene dimension $384$, action-conditioning
dimension $256$, hidden dimension $384$, and dropout $0$.

\noindent\textbf{Safety labels and training.}
Safety supervision is obtained from SafeLIBERO contact-collision auditing. For
each rollout chunk, the dataset stores context latents, the candidate VLA action
chunk, binary contact labels for the valid action steps, and a valid-step mask.
The head is trained with AdamW for $20$ epochs, learning rate $3\times10^{-4}$,
weight decay $1\times10^{-4}$, batch size $64$, and an episode-level train/val
split with validation fraction $0.2$ to avoid chunk leakage across splits. The
loss is binary cross entropy with logits, masked by the valid-step mask. During
evaluation and policy optimization, contact probabilities are obtained by
applying a sigmoid to the contact logits.

\noindent\textbf{Trajectory aggregation.}
The reward and safety heads produce dense per-step signals along an imagined
trajectory. Task reward is averaged over all imagined control steps. Safety cost
uses the top-$M$ aggregation in Eq.~\ref{eq:trajectory_reward_cost} with
$M=16$, emphasizing short high-risk intervals instead of diluting them over the
full horizon. This produces the trajectory-level reward-cost pair used by
safety-constrained GRPO.

\end{document}